\documentclass[letterpaper]{article}
\usepackage{uai2020}
\usepackage[margin=1in]{geometry}

\usepackage{times}

\usepackage{hyperref}
\usepackage{url}

\usepackage{verbatim}
\usepackage[authoryear,round]{natbib}
\usepackage{graphicx}
\usepackage[export]{adjustbox}
\usepackage{subcaption}
\usepackage{amsmath}
\usepackage{amssymb}
\usepackage{wrapfig}
\usepackage{paralist}
\usepackage[usenames,dvipsnames]{xcolor}
\usepackage{tikz}
\usepackage[capitalize,noabbrev]{cleveref} % better autoref (\cref \Cref)
\usepackage[colorinlistoftodos,prependcaption,textsize=small]{todonotes}
\usepackage[acronym,smallcaps,nowarn,section,nogroupskip,nonumberlist]{glossaries}
\usepackage[linesnumbered,ruled,noend]{algorithm2e}

\usetikzlibrary{calc,backgrounds,positioning,bayesnet,decorations.pathmorphing,patterns,fit}
\definecolor{red}{HTML}{E41A1C}
\definecolor{orange}{HTML}{FF7F00}
\definecolor{yellow}{HTML}{FFC020}
\definecolor{green}{HTML}{4DAF4A}
\definecolor{blue}{HTML}{377EB8}
\definecolor{purple}{HTML}{984EA3}
\PassOptionsToPackage{hyphens}{url}
\crefformat{equation}{(#2#1#3)}
\Crefformat{equation}{Equation~#2#1#3}
\crefrangeformat{equation}{Eqs.~#3#1#4 to~#5#2#6}
\Crefrangeformat{equation}{Equations~#3#1#4 to~#5#2#6}
\crefmultiformat{equation}{Eqs.~#2#1#3}{ and~#2#1#3}{, #2#1#3}{ and~#2#1#3}
\Crefmultiformat{equation}{Equations~#2#1#3}{ and~#2#1#3}{, #2#1#3}{ and~#2#1#3}
\glsdisablehyper{}
\newacronym{VAE}{vae}{Aariational Autoencoder}
\newacronym{IWAE}{iwae}{Importance Weighted Autoencoder}
\newacronym{IS}{is}{Importance Sampling}
\newacronym[firstplural=Partially Observable Markov Decision Processes]{POMDP}{pomdp}{Partially Observable Markov Decision Process}
\newacronym[firstplural=Markov Decision Processes]{MDP}{mdp}{Markov Decision Process}
\newacronym{RL}{rl}{Reinforcement Learning}
\newacronym{ADRQN}{adrqn}{Action-specific Deep Recurrent Q-network}
\newacronym{(A)DRQN}{adrqn}{}
\newacronym{RNN}{rnn}{Recurrent Neural Network}
\newacronym{VRNN}{vrnn}{Variational Recurrent Neural Network}
\newacronym{AESMC}{aesmc}{Autoencoding Sequential Monte Carlo}
\newacronym{SMC}{smc}{Sequential Monte Carlo}
\newacronym{DVRL}{dvrl}{Deep Variational Reinforcement Learning}
\newacronym{A3C}{A3C}{Asynchronous Advantage Actor-critic}
\newacronym{A2C}{A2C}{Advantage Actor-critic}
\glsunset{A2C}
\newacronym{ELBO}{elbo}{Evidence Lower Bound}
\newacronym{DQN}{dqn}{Deep Q-network}
\newacronym{DPFP}{dpfp}{Deep Particle Filter Based Policy}
\newacronym{CNN}{cnn}{Convolutional Neural Network}
\newacronym{DRQN}{drqn}{Deep Recurrent Q-network}
\newacronym{KL}{kl}{Kullback-Leibler}
\newacronym{VIN}{vin}{Value Iteration Network}
\newacronym{GRU}{gru}{Gated Recurrent Unit}
\newacronym{LSTM}{lstm}{Long Short Term Mmemory}
\newacronym{NN}{nn}{Neural Network}
\newacronym{SSM}{ssm}{State Space Model}
\newacronym{ADR-A2C}{adr-a2c}{Action-specific Deep Recurrent A$2$C Network}
\newacronym[firstplural=Bayes-adaptive Partically Observable Decision Processes]{BA-POMDP}{ba-pomdp}{Bayes-adaptive Partically Observable Decision Process}
\newacronym{ESS}{ess}{Effective Sample Size}
\newacronym{BPTT}{bptt}{Backpropagation-throught-time}
\newacronym{PAI}{pai}{Planning as Inference}
\newacronym{EM}{em}{expectation maximization}
\newacronym{MLSH}{mlsh}{Meta Learning of Shared Hierarchies}
\newacronym{HRL}{hrl}{Hierarchical RL}
\newacronym{IDOL}{idol}{Inference and Distillation for Option Learning}
\newacronym{MSOL}{msol}{Multitask Soft Option Learning}
\newacronym{OC}{oc}{Option Critic}
\newacronym{Distral}{distral}{}
\newcommand{\Distral}{\textsc{distral}}
\DeclareMathOperator{\E}{\mathbb{E}}

\DeclareMathOperator{\DKL}{{}\mathbb{D}_{\text{\scalebox{0.75}{KL}}}\hspace*{-2pt}}

\newcommand{\KL}[2]{\DKL\left(#1 \,\|\; #2\right)}
\renewcommand{\L}{\ensuremath{\mathcal{L}}}
\renewcommand{\O}{\ensuremath{\mathcal{O}}}

\newcommand{\smallsum}[2]{ {\textstyle
	\sum\limits_{\scriptscriptstyle #1}^{\scriptscriptstyle #2}}
}
\newcommand{\smallprod}[2]{ {\textstyle
	\prod\limits_{\scriptscriptstyle #1}^{\scriptscriptstyle #2}}
}
\newcommand*\circled[1]{\tikz[baseline=(char.base)]{
            \node[shape=circle,draw,inner sep=1pt] (char) {\tiny #1};}}

\title{Multitask Soft Option Learning}

\author{%
  Maximilian Igl\thanks{Corresponding author: \texttt{maximilian.igl@gmail.com}}\\
  University of Oxford\\
  \And
  Andrew Gambardella\\
  University of Oxford\\
  \And
  Jinke He \\
  Delft University of Technology\\
  \And
  Nantas Nardelli\\
  University of Oxford\\
  \AND
  N. Siddharth\\
  University of Oxford\\
  \And
  Wendelin B{\"o}hmer\\
  University of Oxford\\
  \And
  Shimon Whiteson\\
  University of Oxford\\
}

\begin{document}

\maketitle

\begin{abstract}
  We present Multitask Soft Option Learning (\textsc{msol}), a hierarchical multitask framework based on Planning as Inference. 
  \textsc{msol} extends the concept of options, using separate variational posteriors for each task, regularized by a shared prior.
  This ``soft'' version of options avoids several instabilities during training in a
  multitask setting, and provides a natural way to learn both intra-option policies and
  their terminations.
  Furthermore, it allows fine-tuning of options for new tasks without forgetting their learned policies,
  leading to faster training without reducing the expressiveness of the 
  hierarchical policy.
  We demonstrate empirically that \textsc{msol} significantly outperforms both hierarchical and flat transfer-learning baselines.
\end{abstract}

%%% Local Variables:
%%% mode: latex
%%% TeX-master: "../main"
%%% End:

\section{INTRODUCTION}
\vspace{-2mm}

A key challenge in Deep Reinforcement Learning is to scale current approaches
to complex tasks without requiring
a prohibitive number of environmental interactions.
One promising approach is
% For many tasks, it is possible
to construct or learn efficient exploration priors to focus on more relevant
parts of the state-action space, reducing the number of required interactions.
This includes, for example, reward shaping \citep{ng1999policy}, curriculum learning
\citep{bengio2009curriculum}, meta-learning \citep{wang2016learning} and transfer learning \citep{teh2017distral}.
% some meta-learning algorithms \citep{wang2016learning,duan2016rl,gupta2018meta,houthooft2018evolved,pmlr-v80-xu18d}, and
% meta-learning \citep{gupta2018meta}, and
% transfer learning \citep{caruana1997multitask,taylor2011introduction,bengio2012deep,parisotto2015actor,Rusu2015,teh2017distral}.

In particular, transfer learning does not require human designed rewards or curricula, instead
allowing the network to learn what and how to transfer knowledge between tasks.
One promising way to capture such knowledge
is to decompose policies into a hierarchy of sub-policies (or skills) that can be
reused and combined in novel ways to solve new tasks
% \citep{dayan1993feudal,thrun1995finding,parr1998reinforcement,sutton1999between,barto2003recent}.
\citep{sutton1999between}.
This idea of \gls{HRL} is also supported by
findings that humans appear to employ a hierarchical
mental structure when solving tasks
\citep{botvinick2009hierarchically}.
In such a hierarchical policy, lower-level, {\em temporally extended} skills
yield directed behavior over multiple time steps.
This has two advantages:
\begin{inparaenum}[i)]
\item it allows efficient exploration,
	as the target states of skills can be reached without having
	to explore much of the state space in between, and
\item directed behavior also reduces the variance of the future reward,
	which accelerates convergence of estimates thereof.
\end{inparaenum}
On the other hand, while a hierarchical approach
can significantly speed up exploration and training,
it can also severely limit the expressiveness of the final policy and lead to suboptimal performance when the
temporally extended skills are not able to express the required policy for the task at hand.

Many methods exist for learning such hierarchical skills, %e.g.
% \citep{dayan1993feudal,sutton1999between,mcgovern2001automatic,menache2002q,%
% csimcsek2009skill,gregor2016variational,kulkarni2016hierarchical,%
% bacon2017option,nachum2018data}.
\citep[e.g.][]{sutton1999between,bacon2017option,gregor2016variational}.
The key challenge is to learn skills which are diverse, and relevant for future tasks.
One widely used approach is to rely on additional human-designed input, often in the form of
manually specified subgoals \citep{vezhnevets2017feudal,nachum2018data} or a fixed temporal
extension of learned skills \citep{frans2018meta}.
While this can lead to impressive results, it is only applicable in situations where relevant
subgoals or temporal extension can be easily identified a priori.

This paper proposes \gls{MSOL},
an algorithm to learn hierarchical skills
from a given distribution of tasks
without any additional human specified knowledge.
%In this work, we instead rely purely on a given distribution of tasks without any additional human
%specified knowledge.
%The proposed algorithm
\gls{MSOL}
trains simultaneously on multiple tasks
from this distribution and autonomously extracts
sub-policies
% (also called \emph{options} or \emph{skills})
which are reusable across them.

Importantly, unlike prior work \citep{frans2018meta},
our proposed \emph{soft option} framework
avoids several pitfalls of learning options
from multiple tasks, which arise when skills
are jointly optimized with a higher-level
policy that determines when each skill is used.
%This higher-level policy must use each skill 
Generally, as each skill must be used for similar purposes across all tasks,
to learn consistent behavior, a complex training schedules
is required to assure a nearly converged higher-level
policy before skills can be updated \citep{frans2018meta}.
However, once a skill has converged
it can be hard to change its behavior
without hurting the performance of 
higher-level policies that rely it. Training is therefore prone to end up in local optima:
even if changing a skill on \emph{one} task could increase the return, 
it would likely lead to lower returns 
on \emph{other} tasks in which it is currently used.
This is particularly an issue when multiple skills 
have learned similar behavior, preventing the
learning of a diverse set of skills.

%The first problem 
%requires complex training schedules to assure a nearly converged higher-level
%policy before skills can be updated \citep{frans2018meta}.
%The second problem leads to local optima: even if changing a skill on \emph{one} task could increase
%the return, it would likely lead to decreasing returns on \emph{other} tasks in which it is used.
%This is particularly an issue when multiple skills have learned similar behavior, preventing the
%learning of a diverse set of skills.
%We discuss this in further detail in \cref{sec:local_minima}.

\gls{MSOL} alleviates both difficulties.
The core idea is to learn a ``prototypical'' -- or \emph{prior} -- behavior for each skill, while
allowing the actually-executed skill on each task -- the \emph{posterior} -- to deviate from it if the
specific task rewards require it.
Penalizing deviations between the prior and posteriors from different tasks
gives rise to skills that are consistent across tasks,
and can be elegantly formulated in 
the \gls{PAI} framework \citep{levine2018reinforcement}.
This distinction between prior and task-dependent posterior obviates the need for complex training schedules:
every task can change their posterior independently of each other
and discover new skills without direct interference in other tasks.
Nevertheless, the penalization term encourages skills to be similar across tasks \emph{and} rewards
higher-level policies for preferring such more specialised skills.
We discuss in more detail in \cref{sec:local_minima} how this helps to prevent the aforementioned
local optima.

In addition to these optimization pitfalls, 
the idea of soft options also alleviates the 
restrictiveness of hierarchical policies.
New tasks can make use of learned skills,
by initializing their posterior skills from the priors, 
but are not restricted by them.
%Once our skill-priors are learned, we can solve new tasks,
%even if the required policies on those new tasks deviate from our previous skills.
The penalization term between prior and posterior acts here as learned shaping
reward, guiding the exploration on new tasks towards previously relevant behavior, without
requiring the new policy to exactly match previous behavior.
In difference to prior work,
\gls{MSOL} can thus even learn tasks 
that are not solvable with previously learned skills alone.
Finally, we show how the {\em soft option} framework gives rise to a natural solution to the
challenging task of learning option-termination policies.

Our experiments demonstrate that \gls{MSOL}
outperforms previous hierarchical and transfer-learning
algorithms during transfer tasks in a multitask setting.
Unlike prior work, 
\Gls{MSOL} only modifies the regularized reward and loss function.
and does not require specialized architectures, or artificial
restrictions on the expressiveness 
of either the higher-level or intra-option policies.

%%% Local Variables:
%%% mode: latex
%%% TeX-master: "../main"
%%% End:

\vspace*{-1.0ex}
\section{PRELIMINARIES}
\label{sec:math}
\vspace{-2mm}

An agent's task is formalized as
a MDP $(\mathcal S, \mathcal A, \rho, P, r, \gamma)$,
consisting of the state space $\mathcal S$,
action space $\mathcal A$,
initial state distribution $\rho$,
transition probability $P(s_{t+1}|s_t, a_t)$
of reaching state $s_{t+1}$ by executing action~$a_t$ in state $s_t$,
reward $r(s_t, a_t)\in\mathbb{R}$ that an agent receives for this transition, and discount factor $\gamma\in[0,1]$.
An optimal agent chooses actions that maximize the return $R_t(s_t) =
\sum_k \gamma^k r_{t+k}$ consisting of discounted future rewards.

% ---------------------------------------------------------------------------
\subsection{PLANNING AS INFERENCE}
Planning as inference (\gls{PAI})
\citep{todorov2008general,levine2018reinforcement}
frames \gls{RL}
as a probabilistic inference problem.
The agent learns
a distribution $q_\phi(a|s)$
over actions $a$ given states $s$,
i.e., a policy, parameterized by $\phi$,
which induces a distribution over trajectories~\(\tau\) of length $T$,
i.e., $\tau = (s_1, a_1, s_2, \ldots, a_T, s_{T+1})$:
\vspace{-1mm}
\begin{equation}
  \label{eq:variational_posterior}
  q_{\phi}(\tau) =
  \rho(s_1) \smallprod{t=1}{T}	 q_{\phi}(a_t|s_t) \, P(s_{t+1}|s_t,a_t) \,.
\end{equation}
\vspace{-5mm}

This can be seen as a structured variational approximation of the optimal trajectory distribution.
Note that the true initial state probability $\rho(s_1)$ and transition
probability $P(s_{t+1}|s_t,a_t)$ are used in the variational posterior,
as we can only control the policy, not the environment.

A significant advantage of this formulation is that it is straightforward to incorporate
information both from prior knowledge, in the form of a prior policy distribution, and the task at
hand through a likelihood function that is defined in terms of the achieved reward.
The prior policy $p(a_t|s_t)$ can be specified by hand or, as in our case, learned (see \cref{sec:method}).
To incorporate the reward, we introduce a binary
{\em optimality variable} $\mathcal O_t$ \citep{levine2018reinforcement}, whose likelihood is highest along the optimal trajectory that maximizes return:
%
%\begin{equation}
$
  p(\O_t=1|s_t,a_t) =
  	\exp\big(r(s_t,a_t) / \beta\big)
$,
%\end{equation}
%
where for $\beta\to 0$ we recover the original \gls{RL} problem.
The constraint $r \in (-\infty,0]$ can be relaxed
without changing the inference procedure \citep{levine2018reinforcement}.
For brevity, we denote $\mathcal O_t = 1$ as
$\mathcal O_t \equiv (\mathcal O_t = 1)$.
%
%For a given prior policy $p(a_t|s_t)$,
%the distribution of `optimal' trajectories is $p(\!\tau, \O_{1:T}\!)$.
%
%\vspace{-2mm}
%\begin{equation}  \label{eq:variational_target}
%	\hspace{0mm}
%	p(\!\tau, \O_{1:T}\!) =
%	\rho(\!s_1\!) \!\prod_{t=1}^T p(\!a_t|s_t\!) \,
%	P(\!s_{t+1}\!|s_t,a_t\!) \, p(\!\O_t|s_t,a_t\!) . \hspace{-4mm}
%\end{equation}
%
If a given prior policy $p(a_t|s_t)$
explores the state-action space sufficiently, then
$p(\!\tau, \O_{1:T}\!)$ is the distribution of desirable trajectories.
\gls{PAI} aims to find a policy
such that the variational posterior
in \cref{eq:variational_posterior}
approximates this distribution
by minimizing the \gls{KL} divergence:
%
% \vspace{-1mm}
\begin{equation} \label{eq:pai_objective}
  \begin{split}
  \L(\phi) & = \KL{q_\phi(\tau)}{p(\tau, \mathcal O_{1:T})} ,
  \, \text{where} \; \\
  \hspace*{-0.7em}p(\!\tau, \O_{1:T}\!)
  & = \rho(\!s_1\!) \smallprod{t=1}{T} p(\!a_t|s_t\!)
  P(\!s_{t\!+\!1}\!|s_t,a_t\!)  p(\!\O_t|s_t,a_t\!) .
  \end{split}
\end{equation}

% ------------------------------------------------------------------------------
\subsection{MULTI-TASK LEARNING}
In a multi-task setting, we have
a set of different tasks $i\in\mathcal{T}$,
drawn from a task distribution with probability $\xi(i)$.
All tasks share state space $\mathcal S$ and action space $\mathcal A$,
but each task has its own initial-state distribution $\rho_{i}$,
transition probability $P_i(s_{t+1}|s_t, a_t)$,
and reward function $r_i$.
Our goal is to learn $n$ tasks concurrently,
distilling common information
that can be leveraged to learn faster on new tasks from $\mathcal{T}$.
In this setting, the prior policy $p_\theta(a_t|s_t)$ can be learned jointly with the task-specific
posterior policies $q_{\phi_i}(a_t|s_t)$ \citep{teh2017distral}.
To do so, we simply extend \cref{eq:pai_objective} to
% \vspace{-1.5mm}
\begin{equation}
  \begin{split}
  \mathcal{L}(\{\phi_i\},\theta)
  & =\E_{i \sim \xi}\big[
  	\KL{q_{\phi_i}(\tau)}{p_\theta(\tau, \mathcal{O}_{1:T})}\big]  \\
  & = -\frac{1}{\beta} \, \E_{i \sim \xi, \tau \sim q} \left[\sum_{t=1}^T  R^\text{reg}_{i,t} \right],
  \label{eq:multitask-loss}
  \end{split}
\end{equation}
\vspace{-3mm}

where $ R^\text{reg}_{i,t} := r_i(s_t, a_t) -
\beta \ln\frac{q_{\phi_i}(a_t|s_t)}{p_{\theta}(a_t|s_t)}$
is a regularised reward.
Minimizing the loss in \cref{eq:multitask-loss} is equivalent to maximizing the
regularized reward \(R^\text{reg}_{i,t}\).
Moreover, minimizing the term $\E_{\tau \sim q}\!
\big[\ln\frac{q_{\phi_i}(a_t|s_t)}{p_{\theta}(a_t|s_t)}\big]$ implicitly minimizes the expected
\gls{KL}-divergence $\E_{s_t \sim
q}\!\big[\mathbb{D}_\text{KL}[q_{\phi_i}(\cdot|s_t)\|p_\theta(\cdot|s_t)]\big]$.
In practise (see \cref{sec:gradient-updates}) we will also make use of a discount factor $\gamma \in [0,1]$.
For details on how $\gamma$ arises in the \gls{PAI} framework we refer to \citet{levine2018reinforcement}.

% ------------------------------------------------------------------------------
\subsection{OPTIONS}

Options \citep{sutton1999between} are skills that
generalize primitive actions and consist of three components:
\begin{inparaenum}[i)]
\item an intra-option policy~$p(a_t|s_t,z_t)$ that selects primitive actions according to the currently active option~$z_t$,
\item a probability $p(b_t|s_t,z_{t-1})$ of terminating the \emph{previously} active option~\(z_{t-1}\), and
\item an initiation set $\mathcal{I}\subseteq\mathcal{S}$, which we simply assume to be $\mathcal{S}$.
\end{inparaenum}
Note that by construction, the higher-level (or master-) policy $p(z_t| z_{t-1}, s_t, b_t)$ can only select a new option~\(z_t\) if the previous option~\(z_{t-1}\) has terminated.

%%% Local Variables:
%%% mode: latex
%%% TeX-master: "../main"
%%% End:

\vspace*{-1.0ex}
\section{METHOD}
\label{sec:method}
\vspace{-2mm}

% \begin{wrapfigure}{r}{0.4\textwidth}
  % \vspace{-11mm}
  % \hspace*{-1ex}
\begin{figure}
  \centering
  \def\figuresize{6mm}
  \scalebox{1.1}{%
    \begin{tikzpicture}[%
      detm/.style={latent,rectangle,minimum size=\figuresize},
      latm/.style={latent,minimum size=\figuresize},
      bgb/.style={rounded corners=1mm,minimum height=1.5cm,minimum width=2cm},
      ta/.style={>=latex},
      |-/.style={to path={|- (\tikztotarget)}},
      spring/.style={decoration={aspect=0.5,segment length=1.1mm,amplitude=1mm,coil}, decorate},
      thick]
      %% task 1
      % nodes
      \node[obs, minimum size=\figuresize, label=right:\textsc{\bfseries\color{green!90}Task 1}]
      (s1)   {\(s_t^1\)};
      \node[detm, below right=7mm and 1mm of s1] (q1t) {\(q^T_{\phi_1}\)};
      \node[latm, left=2mm of s1] (ztm1)  {\scalebox{0.7}{\(z_{t-1}^1\)}};
      \node[latm, below=3mm of q1t]       (bt1)   {\(b_t^1\)};
      \node[detm, below=4mm of bt1] (q1h) {\(q^H_{\phi_1}\)};
      \node[latm, below=3mm of q1h]       (zt1)   {\(z_t^1\)};
      \node[detm, below=3mm of zt1]       (q1l)   {\(q^L_{\phi_1}\)};
      \node[latm, below=3.8mm of q1l]       (a1)    {\(a_t^1\)};
      % edges
      \edge[ta, |-]                {s1}      {q1h};
      \edge[ta, |-]                {s1|-q1h} {q1l};
      \edge[ta, |-]                {s1|-q1l} {q1t};
      % \edge[ta, bend right=45]     {q1h}     {ztm1};
      % \edge[ta, bend right=45]     {ztm1}    {q1h};
      % \edge[ta, bend right=45]     {ztm1}    {q1t};
      \edge[ta, |-]     {ztm1}    {q1h};
      \edge[ta, |-]     {ztm1}    {q1t};
      \edge[ta]                    {q1t}     {bt1};
      \edge[ta]                    {bt1}     {q1h};
      \edge[ta]                    {q1h}     {zt1};
      \edge[ta]                    {zt1}     {q1l};
      \edge[ta]                    {q1l}     {a1};
      % background
      \scoped[on background layer]
      \node[bgb, fill=green!20, fit={($(q1t.north west)+(-13mm,4mm)$)($(q1l.south east)+(3mm,0mm)$)},
            label={[xshift=0mm,yshift=-5mm]above:\scriptsize\(\)}]
      % \node[bgb, fill=green!20, fit={($(q1t.north east)+(7.5mm,4mm)$)($(q1l.south west)-(5mm,0mm)$)},
      %       label={[xshift=0mm,yshift=-5mm]above:\scriptsize\(q_{\phi_1}(a_t^1 | s_t^1)\)}]
      (st) {};
      %
      %% task 2
      % nodes
      \node[obs, minimum size=\figuresize, right=3cm of s1, label=left:\textsc{\bfseries\color{blue!90}Task 2}]
      (s2) {\(s_t^2\)};
      \node[detm, below left=7mm and 1mm of s2] (q2t) {\(q^T_{\phi_2}\)};
      \node[latm, right=2mm of s2] (ztm2)  {\scalebox{0.7}{\(z_{t-1}^2\)}};
      \node[latm, below=3mm of q2t]       (bt2)   {\(b_t^2\)};
      \node[detm, below=4mm of bt2] (q2h) {\(q^H_{\phi_2}\)};
      \node[latm, below=3mm of q2h]      (zt2)   {\(z_t^2\)};
      \node[detm, below=3mm of zt2]      (q2l)   {\(q^L_{\phi_2}\)};
      \node[latm, below=3.8mm of q2l]      (a2)    {\(a_t^2\)};
      % edges
      \edge[ta, |-]                {s2}      {q2h};
      \edge[ta, |-]                {s2|-q2h} {q2l};
      \edge[ta, |-]                {s2|-q2l} {q2t};
      % \edge[ta, bend left=45]      {q2h}     {ztm2};
      % \edge[ta, bend left=45]      {ztm2}    {q2h};
      % \edge[ta, bend left=45]      {ztm2}    {q2t};
      \edge[ta, |-]                {ztm2}    {q2h};
      \edge[ta, |-]                {ztm2}    {q2t};
      \edge[ta]                    {q2t}     {bt2};
      \edge[ta]                    {bt2}     {q2h};
      \edge[ta]                    {q2h}     {zt2};
      \edge[ta]                    {zt2}     {q2l};
      \edge[ta]                    {q2l}     {a2};
      % background
      \scoped[on background layer]
      \node[bgb, fill=blue!20, fit={($(q2t.north east)+(13mm,4mm)$)($(q2l.south west)-(3mm,0mm)$)},
            label={[xshift=0mm,yshift=-5mm]above:\scriptsize\(\)}]
      % \node[bgb, fill=blue!20, fit={($(q2t.north west)+(-7.5mm,4mm)$)($(q2l.south east)-(-5mm,0mm)$)},
      %       label={[xshift=0mm,yshift=-5mm]above:\scriptsize\(q_{\phi_2}(a_t^2 | s_t^2)\)}]
      (st) {};
      %
      %% connect with priors
      \node[detm, draw=BrickRed, right=0.56cm of q1t.0,
            label=above:\textsc{\bfseries\color{BrickRed}\,Priors}] (pt) {\(p^T_\theta\)};
      \node[detm, draw=BrickRed, right=0.56cm of q1h.0] (ph) {\(p^H\)};
      \node[detm, draw=BrickRed, right=0.56cm of q1l] (pl) {\(p^L_{\theta}\)};
      \edge[-, spring, BrickRed] {q1t.0}  {pt};
      \edge[-, spring, BrickRed] {pt}      {q2t.180};
      \edge[-, spring, BrickRed] {q1h.0} {ph};
      \edge[-, spring, BrickRed] {ph}      {q2h.180};
      \edge[-, spring, BrickRed] {q1l}     {pl};
      \edge[-, spring, BrickRed] {pl}      {q2l};
    \end{tikzpicture}}
  % \vspace{-5mm}
  \caption{\small
  		Two hierarchical posterior policies (left and right)
  		with common priors (middle).
  		For each task $i$, the policy conditions on
  		the current state $s_t^i$ and the last selected option $z_{t-1}^i$.
  		It samples, in order, whether to terminate the last option ($b_t^i$),
  		which option to execute next
  		($z_t^i$) and what primitive action ($a_t^i$)
  		to execute in the environment.}
  \label{fig:schema}
  \vspace*{-2ex}
\end{figure}
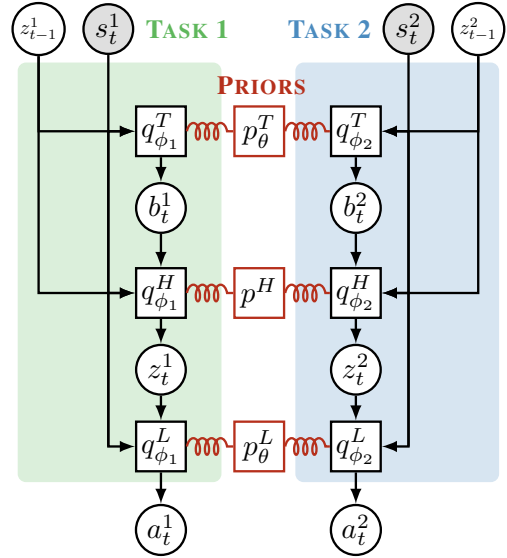
  % \vspace{-19mm}
% \end{wrapfigure}
%

We aim to learn a reusable set of options
that allow for faster
training on new tasks from a given distribution.
To differentiate ourselves from classical `hard' options,
which, once learned, do not change during new tasks,
we call our novel approach {\em soft-options}.
% (this is further discussed in
% \cref{sec:relationship}).
Each soft-option consists of an option {\em prior}, denoted by $p_\theta$,
which is shared across all tasks,
and a task-specific option {\em posterior}, denoted by $q_{\phi_i}$ for task $i$.
Unlike most previous work, e.g. \citep{frans2018meta}, we learn both intra-option and termination policies.
The priors of both the intra-option policy $p_\theta^L$ and the termination policy $p_\theta^T$ capture how an option
typically behaves and remain fixed once they are fully learned.
At the beginning of training on a new task, they are used to initialize the
task-specific posterior distributions $q_{\phi_i}^L$ and $q_{\phi_i}^T$.
During training, the posterior is then regularized against the
prior to prevent inadvertent unlearning. However, if maximizing the reward on certain
tasks is not achievable with the prior policy, the posterior is free to deviate from it.
We can thus speed up training using options, while remaining flexible enough to solve more tasks.
Additionally, this soft option framework also allows for learning good {\em priors} in a multitask setting while
avoiding complex training schedules and local optima (see \cref{sec:local_minima}).
In this work, we also learn the higher-level posterior $q_{\phi_i}^H$ within the framework of
\gls{PAI}, but assume a fixed, uniform prior distribution $p^H$, i.e. we assume there is no shared higher-level structure
between tasks.
\Cref{fig:schema} shows an overview over this architecture which we explain further below.

% ------------------------------------------------------------------------------
\subsection{HIERARCHICAL POSTERIOR POLICIES}

To express options in the \gls{PAI} framework,
we introduce two additional variables at each time step $t$:
{\em option selections} $z_t$, representing the currently selected option,
and decisions $b_t$ to {\em terminate}\
them and allow the higher-level (master) policy to choose a new option.
The agent's behavior depends on the currently selected option $z_t$,
by drawing actions $a_t$ from the
{\em intra-option posterior policy} $q^L_{\phi_i}(a_t|s_t, z_t)$.
The selection $z_t$ itself is drawn from a {\em master policy}
%$q^H_{\phi_i}(z_t | s_t, z_{t-1}, b_t)$.
$q^H_{\phi_i}(z_t | s_t, z_{t-1}, b_t) =
	(1 - b_t) \, \delta(z_t - z_{t-1})
	\;+\; b_t \, q^H_{\phi_i}(z_t | s_t)$,
which conditions on $b_t \in \{0,1\}$,
drawn by the {\em termination posterior policy}
$q^T_{\phi_i}(b_t|s_t, z_{t-1})$.
The master policy either continues with the previous $z_{t-1}$
or draws a new option,
where
we set $b_1=1$ at the beginning of each episode.
We slightly abuse notation by referring by $\delta(z_t-z_{t-1})$ to the Kronecker delta
$\delta_{z_t,z_{t-1}}$ for discrete and the Dirac delta distribution for continuous $z_t$.
The joint posterior policy is
\begin{equation}
	\label{eq:policy_posterior}
	\begin{split}
	& q_{\phi_i}(a_t, z_t, b_t | s_t, z_{t-1})  = \\
		&q^T_{\phi_i}(b_t|s_t, z_{t-1}) \,
		q^H_{\phi_i} (z_t|s_t, z_{t-1}, b_t) \,
		q^L_{\phi_i}(a_t|s_t, z_t) \,.
	\end{split}
\end{equation}
While $z_t$ can be a continuous variable,
we consider only $z_t\in \{1 \dots m\}$,
where $m$ is the number of available options.
The induced distribution $q_{\phi_i}(\tau)$ over trajectories of task $i$,
$\tau = (s_1, b_1, z_1, a_1, s_2, \ldots, s_T, b_T, z_T, a_T, s_{T+1})$,
is then
% \vspace{-.5mm}
\begin{equation} \label{eq:joint_posterior}
	\begin{split}
	\textstyle
	 q_{\phi_i}(\tau)\! =\! \rho_{i}(s_1) \smallprod{t=1}{T}
		q_{\phi_i}\!(a_t, \!z_t, \!b_t|s_t, \!z_{t\!-\!1})
		P_i(\!s_{t\!+\!1}|s_t, \!a_t\!) . \!\!
	\end{split}
\end{equation}
% \vspace{-3mm}

% ------------------------------------------------------------------------------
\subsection{HIERARCHICAL PRIOR POLICY}
Our framework transfers knowledge between tasks
by a shared prior $p_\theta(\!a_t, \!z_t, \!b_t|s_t, \!z_{t-\!1}\!)$
over all joint policies \eqref{eq:policy_posterior}:
\hspace{-2mm} % Because the line above is juuuust a bit too long
\begin{equation}
  \begin{split}
& p_\theta(a_t,z_t, b_t|s_t, z_{t-1})  =  \\
  		& p_\theta^T\!(b_t|s_t, z_{t-1}) \,
  		p^H\!(z_t|z_{\scriptscriptstyle t-1}, b_t) \,
  		p_{\theta}^L\!(a_t|s_t,z_t).
  \end{split}
\end{equation}
By choosing $p^T_\theta$, $p^H$, and $p^L_\theta$ correctly,
we can learn useful temporally
extended options.
The parameterized priors $p_\theta^T(b_t|s_t, z_{t-1})$ and $p_{\theta}^L(a_t|s_t,z_t)$ are structurally
equivalent to the posterior policies $q^T_{\phi_i}$
and $q^L_{\phi_i}$ so that they can be used as
initialization for the latter on new tasks.
Optimizing the regularized return (see next section)
w.r.t.~$\theta$ distills the common behavior
into the prior policy and softly enforces similarity across
posterior distributions of each option amongst all tasks $i$.

The prior $p^H(z_t | z_{t-1}, b_t)
= (1 - b_t) \, \delta(z_t - z_{t-1})
\,+\, b_t {\textstyle\frac{1}{m}}$
selects the previous option $z_{t-1}$
if $b_t=0$, and otherwise draws options uniformly to ensure exploration.
Because the posterior master policy is different on each task, there is no need to distill
common behavior into a joint prior.

% ------------------------------------------------------------------------------
\subsection{OBJECTIVE}
\label{sec:objective}
We extend the multitask objective in \eqref{eq:multitask-loss}
by substituting $p_\theta(\tau, \mathcal{O}_{1:T})$ and $p_{\phi_i}(\tau)$
with those induced by our hierarchical
posterior policy in \eqref{eq:policy_posterior}
and the corresponding prior.
The resulting objective has the same form
but with a new regularized reward that is maximized:
%instead of \cref{eq:multitask-reward}:
%\vspace{-1mm}
\begin{equation}
  \label{eq:full_reg_reward}
  \begin{split}
  R_{i,t}^\text{reg}
   = &  r_i(s_t,a_t)
  	- \underbrace{\beta \ln{\textstyle\frac{q^H_{\phi_i}\!(z_t|s_t, z_{t-1}, b_t)}%
  			{p^H\!(z_t|z_{t-1}, b_t)}}}_{\circled{1}} \\[-2mm]
	& - \underbrace{\beta \ln{\textstyle\frac{q^L_{\phi_i}(a_t|s_t, z_t)}%
			{p^L_\theta\!(a_t|s_t, z_t)}}}_{\circled{2}}
	- \underbrace{\beta \ln{\textstyle\frac{q^T_{\phi_i}\!(b_t|s_t, z_{t-1})}%
			{p_\theta^T\!(b_t|s_t,z_{t-1})}}}_{\circled{3}} \,.
  \end{split}
\end{equation}
As we maximize $\mathbb{E}_q[R^{\text{reg}}_{i,t}]$,
this corresponds to maximizing the expectation over
\begin{equation}
	r_i(\!s_t,\!a_t\!)
	- \beta \big[\mathbb{D}_\text{\sc kl}(q^H_{\phi_i}\!\| p^H)
		+ \mathbb{D}_\text{\sc kl}(q^L_{\phi_i} \!\| p_\theta^L )
		+ \mathbb{D}_\text{\sc kl}(q_{\phi_i}^T \!\| p^T_\theta) \big],
\end{equation}
along the on-policy trajectories drawn from $q_{\phi_i}(\tau)$.
In the following, we will discuss the effects of all three regularization terms on the optimization.

Term \circled{1} of the regularization encourages exploration in the space of options since we chose
a uniform prior for $p^H$ when the previous option was terminated. It can {\em also} be seen as a form
of deliberation cost \citep{harb2017waiting} as it is only nonzero whenever we
terminate an option and the master policy needs to select another to execute: if the option is not
terminated, we have $z_t=z_{t-1}$ with probability $1$ for both prior and posterior by construction
and $\mathbb{D}_\text{\sc kl}(q^H_{\phi_i}\!\| p^H)=0$. 

Because \cref{eq:full_reg_reward} is optimized across \emph{all} tasks $i$, term \circled{2} updates the prior
towards the `average' posterior. It also regularizes each posterior towards this prior.
This enforces similarity between option posteriors across tasks.
Importantly, it also encourages the \emph{master} policy to pick the most
\emph{specialized} option that still maximizes the return, i.e the option for which
the posteriors $q_{\phi_i}^L$ are most similar across tasks as this will minimize term \circled{2}.
Consequently, if multiple options have learned the desired behavior, the master policy will 
only pick the most specialized option consistently.
As discussed %in more detail
in \cref{sec:local_minima}, this allows us to escape the local
optima that hard options face in multitask learning, while still having fully
specialized options after training.
% This creates the necessary coordination between master policies that allows us to escape the local
% minimum described in \cref{fig:2balls}: if two options $z_1$ and $z_2$ can be used for subgoal $A$
% on task $(a)$, but the option posterior of $z_2$ also learns to reach $B$ on task $(b)$, the master policy on task
% $(a)$ starts to use only $z_1$ (because it is more specialized), leaving $z_2$ to specialize on
% $B$.
% Consequently, despite only softly regularizing the option posteriors towards their joint prior, this
% term leads to specialized options after training as long as the number of available options is sufficient.

Lastly, we can use \circled{3} to also encourage temporal abstraction of options.
To do so, \emph{during option learning}, we fix the termination prior $p^T$ to a Bernoulli distribution
$p^T(b) = (1-\alpha)^b \alpha^{1-b}$.
Choosing a large $\alpha$ encourages prolonged execution
of one option, but allows switching whenever necessary.
This is similar to deliberation costs \citep{harb2017waiting} but with a more flexible cost model.

We can still distill a termination prior $p^T_\theta$ which can be used on future tasks.
Instead of learning $p^T_\theta$ by minimizing the KL against the posterior termination
policies, we can get more decisive terminations by minimizing
\begin{equation}
	\hspace{-0mm}
	\min_\theta  \; \smallsum{i=1}{n} \, \mathbb{E}_{\tau \sim q^i} \! \left[
	\KL{\hat q_{\phi_i}(\cdot|s_t, \!z_{t-1})\!}%
		{\!p_\theta^T(\cdot|s_{t}, \!z_{t-1})} \right]\!,  \hspace{-2mm}
\end{equation}
and
$\hat q_{\phi_i}\!(b\!=\!1|s_t, z_{t-1})
\!\!=\!\! \sum_{z_t\!\neq\! z_{t-1}} \!q^H_{\phi_i}(z_t |s_t, z_{t-1},b_t\!=\!1)$
i.e., the learned termination prior distills
the probability that the tasks' master policies
would change the active option if they had the opportunity.
Details on how we optimized the MSOL objective are given
in \cref{sec:training_details}.

% ------------------------------------------------------------------------------
\subsection{MSOL VS.\ CLASSICAL OPTIONS}
\label{sec:relationship}
Assume we are faced with a new task and are given some prior knowledge in the form of a set
of skills that we can use.
Using the skills' policies and termination probabilities 
as prior policies $p^T$ and $p^L$ in the soft option framework, 
we can interpret $\beta$ as a
temperature parameter determining how closely we are required to follow them. 
For $\beta\to\infty$ we recover the classical ``hard'' option case and our posterior option
policies are restricted to the prior.\footnote{However, in this limiting case optimization
using the regularized reward is not possible.}
For $\beta=0$ the priors only initialize the otherwise unconstrained policy,
quickly unlearning behavior that may be useful down the line.
Only for $0<\beta<\infty$ 
%we softly restrict ourselves to the given skills, which allows us to 
\gls{MSOL} can keep prior information to guide long-term exploration but
can also explore policies ``close'' to them.

% ------------------------------------------------------------------------------
\subsection{LOCAL OPTIMA OPTION LEARNING}
\label{sec:local_minima}

% \begin{wrapfigure}{r}{0.6\linewidth}
\begin{figure}
  \centering
  \begin{subfigure}[b]{0.32\linewidth}
    \centering
    \includegraphics[width=0.85\linewidth]{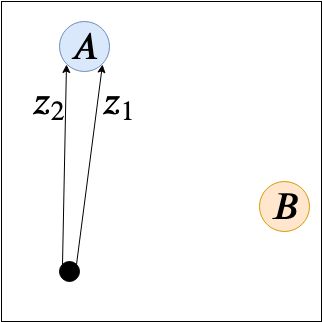}
    \caption{\small Hard options}\vspace*{-1.5ex}
    \label{fig:2balls-ol}
  \end{subfigure}
  \begin{subfigure}[b]{0.32\linewidth}
    \centering
    \includegraphics[width=0.85\linewidth]{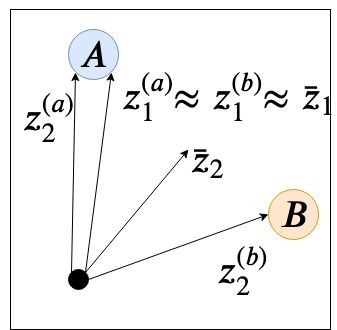}
    \caption{\small Soft options}\vspace*{-1.5ex}
    \label{fig:2balls-pp}
  \end{subfigure}
  \begin{subfigure}[b]{0.32\linewidth}
    \centering
    \includegraphics[width=0.85\linewidth]{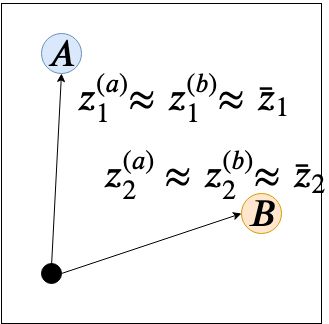}
    \caption{\small After training}\vspace*{-1.5ex}
    \label{fig:2balls-pp2}
  \end{subfigure}
  \vspace{0mm}
  \caption{\small%
    Hierarchical learning of two concurrent tasks ($a$ and $b$) using
    two options ($z_1$ and $z_2$) to reach two relevant targets ($A$ and $B$).
    a) Local optimum when simply sharing options across tasks.
    b) Escaping the local optimum by using prior ($\bar{z}_i$) and posterior ($z_i^{(j)}$) policies.
    c) Learned options after training.
    Details are given in the text in \cref{sec:local_minima}.
  }
  \label{fig:2balls}
% \end{wrapfigure}
\end{figure}

In this section our aim is to provide an intuitive explanation of why learning hard options in a multitask
setting can lead to local optima and how soft options can overcome this.
In this local optimum, multiple options have learned the same behavior and are unable to change it,
even if doing so would ultimately lead to a higher reward.
We use the Moving Bandits experiment schematically depicted in \cref{fig:2balls} as an example. The
agent (black dot) observes two target locations $A$ and $B$ but does not know which one is the
correct one that has to be reached in order to generate a reward.
The state- and action-spaces are continuous, requiring multiple actions to reach either $A$ or $B$
from the starting position. Consequently, having access to two options, one for each location, can
accelerate learning.
Experimental results comparing \gls{MSOL} against a recently proposed `hard option' method
(\gls{MLSH}, \citep{frans2018meta}) are discussed in \cref{sec:moving-bandits}.

Let us denote the options we are learning as $z_1$ and $z_2$ and further assume that due to random
initialization or late discovery of target $B$, both skills currently reach $A$.
% Furthermore, assume that we are training on multiple task \emph{simultaneously} as often the case in
% \gls{RL}.
In this situation, the master policies on tasks in which the correct goal is $A$ are indifferent
between using $z_1$ and $z_2$ and will consequently use \emph{both} with equal probability.

In the case of hard options, changing one skill, e.g.~$z_2$, towards $B$ in order to solve tasks in which $B$ is the
correct target, decreases the performance on all tasks that currently use $z_2$ to reach target $A$,
because for hard options the skills are shared exactly across tasks.
Averaged across all tasks, this would at first \emph{decrease} the overall average return,
preventing any option from changing away from $A$, leaving $B$ unreachable and training stuck in a
local optimum.

To ``free up'' $z_2$ and learn a new skill reaching $B$,
all master policies need to refrain from using $z_2$ to reach $A$
and instead use the equally useful skill $z_1$ exclusively.
Importantly, using soft options makes this possible.
In \Cref{fig:2balls-pp,fig:2balls-pp2} we depict this schematically. The key difference is that in \gls{MSOL} we have separate
\emph{task-specific posteriors} $z_i^{(a)}$ and $z_i^{(b)}$ for tasks $a$ and $b$ and soft options
$i\in\{1,2\}$ (for simplicity, we assume that the correct target is $A$ for task $a$ and $B$ for task $b$).
This allows us, in a first step, to solve \emph{all} tasks (\cref{fig:2balls-pp}):
despite master policies on tasks $a$ still using posterior $z_2^{(a)}$ to reach $A$, the other
posterior $z_2^{(b)}$ can learn to reach $B$.
However, this now makes option $z_2$ less specialized across tasks, i.e. the prior
$\bar{z}_2$ does not agree with either posterior $z_2^{(a)/(b)}$.
Consequently, for tasks $a$, the master policies will now strictly prefer option $z_1$ to reach $A$,
allowing option $z_2$ to specialize on only reaching $B$, leading to the situation shown in
\cref{fig:2balls-pp2} in which both options specialize to reach different targets.
% . Later, as discussed in \cref{sec:objective}, maximizing the regularized reward
% \cref{eq:full_reg_reward} will lead to specialized skills, i.e. option \emph{priors} (\cref{fig:2balls-pp2}).

%%% Local Variables:
%%% mode: latex
%%% TeX-master: "../main"
%%% End:

\vspace*{-1.0ex}
\section{RELATED WORK}
\vspace{-2mm}

Most hierarchical approaches rely on proxy rewards to train the lower level components and their terminations.
Some of them aim to reach pre-specified subgoals \citep{sutton1999between},
which are often found by analyzing the structure of the MDP 
\citep{mcgovern2001automatic}, previously
learned policies \citep{tessler2017deep} or predictability \citep{harutyunyan2019termination}.
Those methods typically require knowledge, or a sufficient approximation, of the transition
model, both of which are often infeasible.

Recently, several authors have proposed unsupervised training objectives for learning diverse skills based on their distinctiveness
\citep{gregor2016variational}.
However, those approaches don't learn termination functions and cannot 
guarantee that the required behavior on the downstream task is included in the set of learned
skills.
\citet{hausman2018learning} also incorporate reward information, but do not learn termination
policies and are therefore restricted to learning multiple solutions to the provided task instead of
learning a {\em decomposition} of the task solutions which can be re-composed to solve new tasks. 
% Their off-policy training algorithm, based
% on Retrace \citep{munos2016safe} and SVG \citep{heess2015learning}, extends straightforwardly
% to our setting.

A third usage of proxy rewards is by training lower level policies to move towards goals defined by the
higher levels. 
When those goals are set in the original state space \citep{nachum2018data}, this approach has difficulty scaling to high
dimensional state spaces like images. Setting the goals in a learned embedding space
\citep{vezhnevets2017feudal} can be difficult to train, though. In both cases,
the temporal extension of the learned skills are set manually. On the other hand,
\citet{goyal2018transfer} also learn a hierarchical agent, but not to transfer skills, but to find
decisions states based on how much information is encoded in the latent layer.

% In this work, we do not employ any proxy reward functions for the lower level policy.
% Instead, we are training the entire hierarchy on the same regularized reward, 
% which guarantees that the learned skills are useful on the task distribution at hand. 
% This is similar to the Option-Critic framework \citep{bacon2017option,smith2018inference}, however,
% we are using a multitask setting which allows us to learn options that generalize better to
% a pre-specified range of tasks \citep{thrun1995finding,frans2018meta}.

HiREPS \cite{daniel2012hierarchical} 
also take an inference motivated approach to learning options. 
In particular \citet{daniel2016probabilistic}
propose a similarly structured hierarchical policy, albeit in a single task setting. 
However, they do not utilize learned prior \emph{and} posterior distributions, but instead use expectation
maximization to iteratively infer a hierarchical policy to explain the current
reward-weighted trajectory distribution. 
% This learns a model, similar to the prior in our framework, 
% but without the additional flexibility provided by our
% learned posterior distribution.  

Several previous works try to overcome the restrictive nature of options that can lead to
sub-optimal solutions by allowing the higher-level actions to modulate the behavior of the
lower-level policies \citet{heess2016learning,haarnoja2018latent}. 
However, this significantly increases the required complexity of the higher-level policy and
therefore the learning time. 
%especially compared to soft options in situations where the learned
%prior sub-policies already allow for a near-optimal solution. 

The multitask- and transfer-learning setup used in this work is inspired by
\citet{thrun1995finding} who suggests extracting options by using
commonalities between solutions to multiple tasks.
Prior multitask approaches often rely on additional human supervision like policy sketches
\citep{andreas2017modular} or desirable sub-goals
\citep{tessler2017deep} in order to learn skills which
transfer well between tasks. 
In contrast, our work aims at finding good termination states without such supervision.
\citet{tirumala2019exploiting} investigate the use of different priors for the higher-level policy
while we are focussing on learning transferrable option priors.
Closest to our work is \gls{MLSH} \citep{frans2018meta} which, however, shares the lower-level
policies across all tasks without distinguishing between prior and posterior and does not learn
termination policies.
As discussed, this leads to local minima and insufficient diversity in the learned options.
Similarly to us, \citet{fox2016principled} differentiate between prior and posterior policies on
multiple tasks and utilize a KL-divergence between them for training. However, they do not consider
termination probabilities and instead only choose one option per task.
% Instead of transferring option policies between tasks, \citet{ammar2014online} aim to share behavior
% through a latent embedding. 
% Another interesting approach to multitask learning is
% \citep{mankowitz2016adaptive} which learns decision regions that are linear in the state instead of
% learning nonlinear master- and termination policies.

Our approach is closely related to \Distral~\citep{teh2017distral} with which we share the 
multitask learning of prior and posterior policies.
However, \Distral~has no hierarchical structure and applies the same prior distribution over
primitive actions, independent of the task. As a necessary hierarchical heuristic, the authors
propose to also condition on the last primitive action taken.
This works well when the last action is indicative of future behavior; however, in
\cref{sec:experiments} we show several failure cases where a {\em learned} hierarchy is needed.

\vspace*{-1.0ex}
\section{EXPERIMENTS}
\vspace{-2mm}
\label{sec:experiments}
\begin{figure*}[ht!]
  \centering
  \begin{subfigure}[b]{0.32\textwidth}
    \includegraphics[width=1\linewidth]{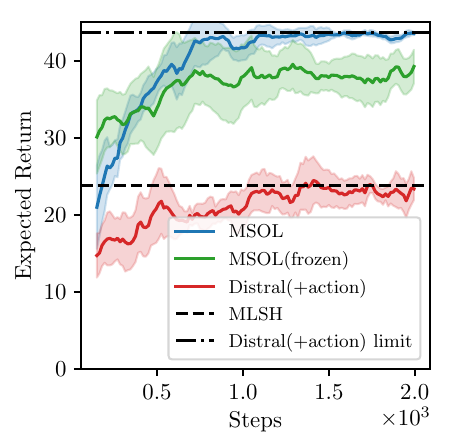}
    % \vspace*{-6mm}
    \caption{\small Moving Bandits}%\vspace*{-1.5ex}
    \label{fig:mv}
  \end{subfigure}
  \begin{subfigure}[b]{0.32\textwidth}
    \includegraphics[width=1\linewidth]{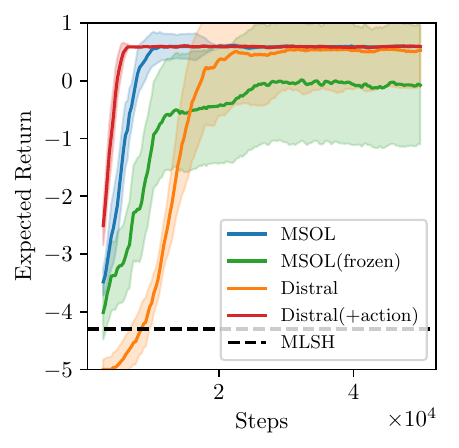}
    % \vspace*{-6mm}
    \caption{\small Taxi}%\vspace*{-1.5ex}
    \label{fig:tn}
  \end{subfigure}
  \begin{subfigure}[b]{0.32\textwidth}
    \includegraphics[width=1\linewidth]{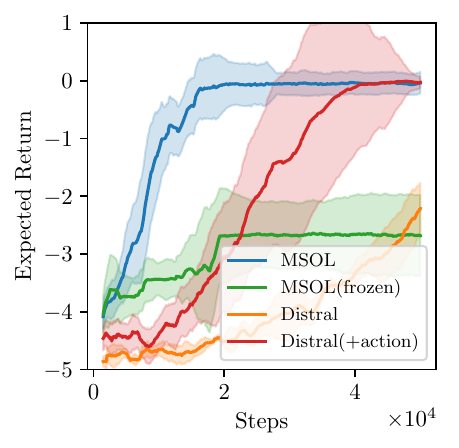}
    % \vspace*{-6mm}
    \caption{\small Directional Taxi}%\vspace*{-1.5ex}
    \label{fig:ts}
  \end{subfigure}
    \vspace*{-1ex}
    \caption{\small
        Performance of applying the learned options and exploration priors to new tasks.
        Each line is the median over 5 random seeds (2 for MLSH) and shaded areas indicated standard
        deviations. 
        Performance during the training phase is shown in \cref{fig:training}.
        \emph{Moving Bandits} (a) is a simple environment capturing the effects described in
        \cref{sec:local_minima}. The results show that \gls{MLSH}, which uses hard options, struggles with
        local minima during the learning phase, whereas \gls{MSOL} is able to learn a diverse set of
        options.
        \emph{Taxi} (b) and \emph{Directional Taxi} (c) additionally require good termination
        policies, which \gls{MLSH} cannot learn as it uses a fixed option duration.
        See \cref{fig:taxi} for a visualization of the options and terminations learned by \gls{MSOL}.
        \Distral(+action) is a strong non-hierarchical baseline which uses the last action as option-heuristic,
        but suffers when that action is not very informative, for example in (c).
      }
  \label{fig:performance}
  % \vspace{-4mm}
\end{figure*}

We conduct a series of experiments to show:
i) \gls{MSOL} trains successfully without complex training schedules like in \gls{MLSH} \citep{frans2018meta},
ii) \gls{MSOL} can learn useful termination policies,
iii) when learning hierarchies in a multitask setting, unlike other methods, \gls{MSOL} successfully overcomes the local
minimum of insufficient option diversity, as described in \cref{sec:local_minima},
iv) using soft options yields fast transfer learning while still reaching optimal performance,
even on new, out-of-distribution tasks.

All architectural details and hyper-parameters 
can be found in the appendix.
For all experiments, 
we first train the exploration priors and options on $n$ tasks 
from the available task distribution $\mathcal{T}$ 
(training phase is plotted in Appendix \ref{sec:hyper-params}). 
Subsequently, 
we test how quickly we can learn new tasks 
from $\mathcal{T}$ (or another distribution $\mathcal{T}'$).

We compare the following algorithms: 
\gls{MSOL} is our proposed method that utilizes soft options
both during option learning and transfer. 
\gls{MSOL}(frozen) uses the soft options framework during learning to find more diverse skills,
but does not allow fine-tuning the posterior sub-policies after transfer. 
% The difference between the two algorithms 
% shows the advantage of the adaptability of soft options for transfer to new tasks.
\Distral~\citep{teh2017distral} is a strong 
non-hierarchical transfer learning algorithm 
that also utilizes prior and posterior distributions. 
\Distral(+action) utilizes the last action as option-heuristic, that is, as additional input to the
policy and prior,
which works well in some tasks but fails when the last action 
is not sufficiently informative.
Conditioning on an \emph{informative} last action allows the \Distral~prior to learn temporally correlated exploration
strategies.
\gls{MLSH} \citep{frans2018meta} 
is a multitask option learning algorithm like \gls{MSOL}, 
but utilizes `hard' options for both learning and transfer, i.e., 
sub-policies that are shared exactly across tasks.
It relies on fixed option durations and requires a complex training schedule between master and
intra-option policies to stabilize training. We use the author's \gls{MLSH} implementation.
We also compare against \gls{OC} \citep{bacon2017option}, which takes the task-id as additional
input in order to apply it to multiple tasks.

Note that, during test time, \gls{MLSH} and \gls{MSOL}(frozen) can be fairly compared as each uses
one fixed policy per skill.  
On the other hand, \Distral, \Distral(+action) and \gls{MSOL} use adaptive posterior policies for
each task and are consequently more expressive.

% ------------------------------------------------------------------------------
\subsection{MOVING BANDITS}
\label{sec:moving-bandits}
\begin{figure*}[ht!]
\vspace{-3mm}
\begin{subfigure}[b]{\linewidth}%{0.9\textwidth}
   \includegraphics[width=.495\linewidth]{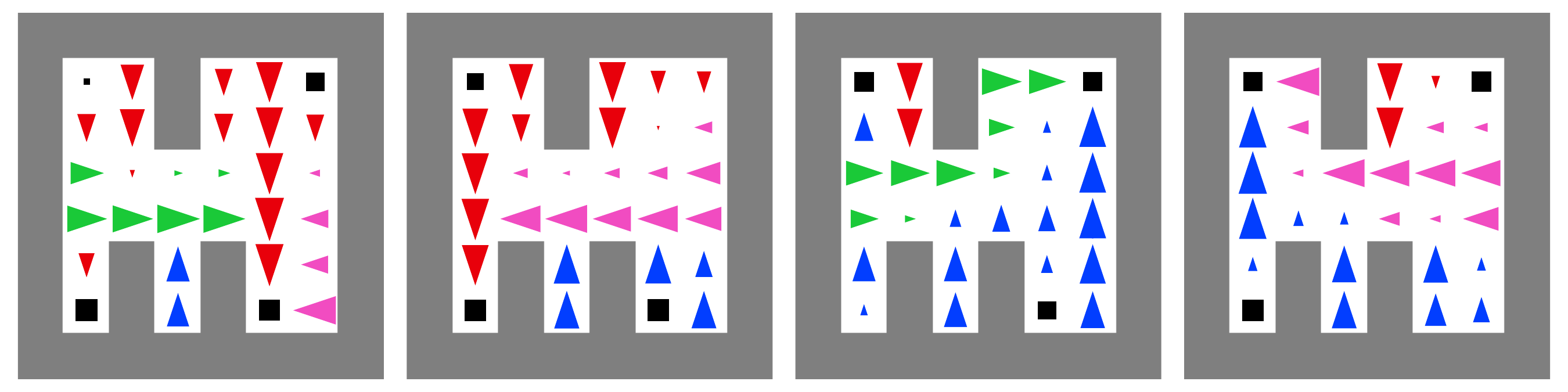}
   \hfill
   \includegraphics[width=.495\linewidth]{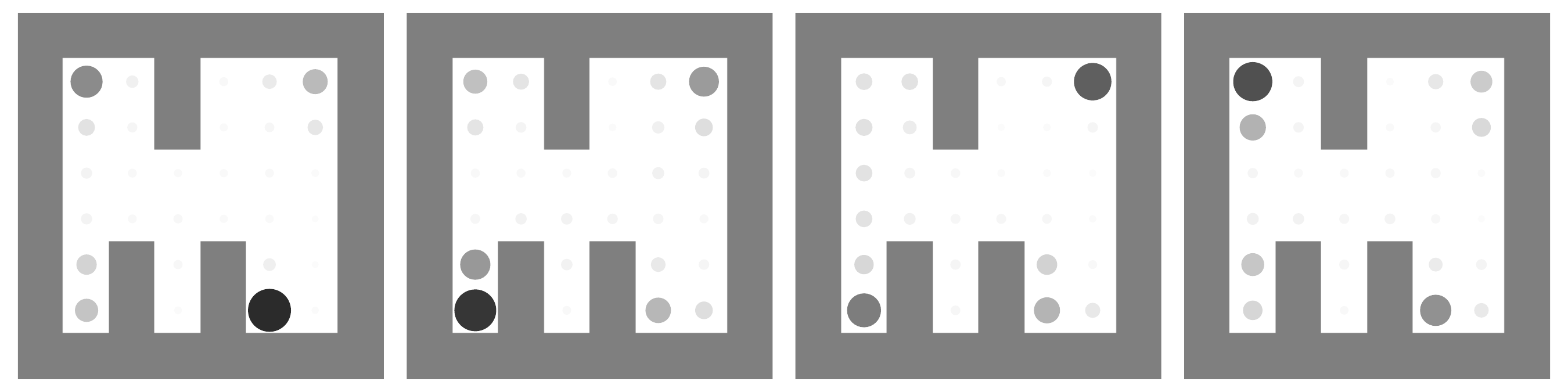}
\end{subfigure}

\begin{subfigure}[b]{\linewidth}%{0.9\textwidth}
	\includegraphics[width=.495\linewidth]{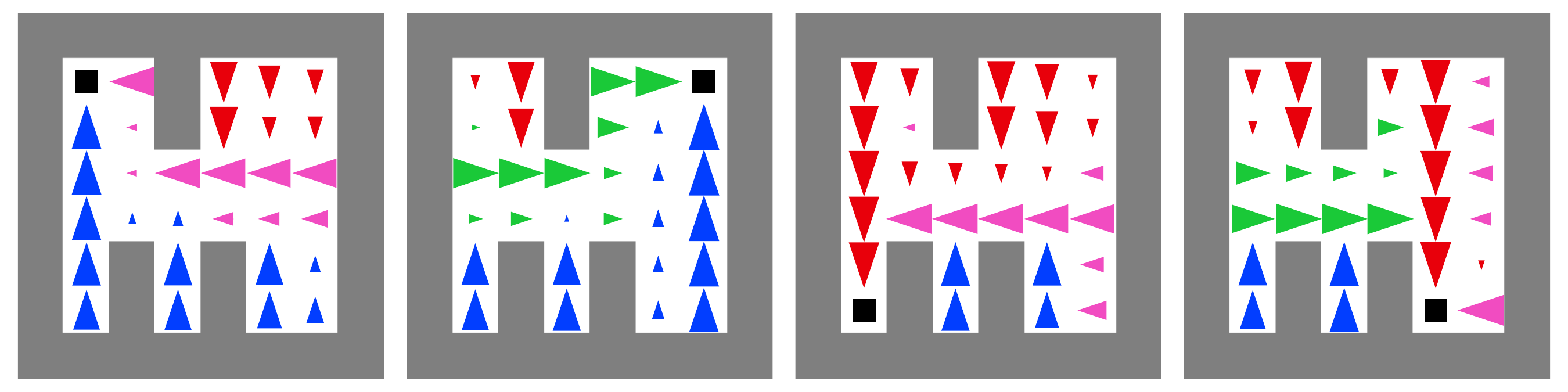}
	\hfill
	\includegraphics[width=.495\linewidth]{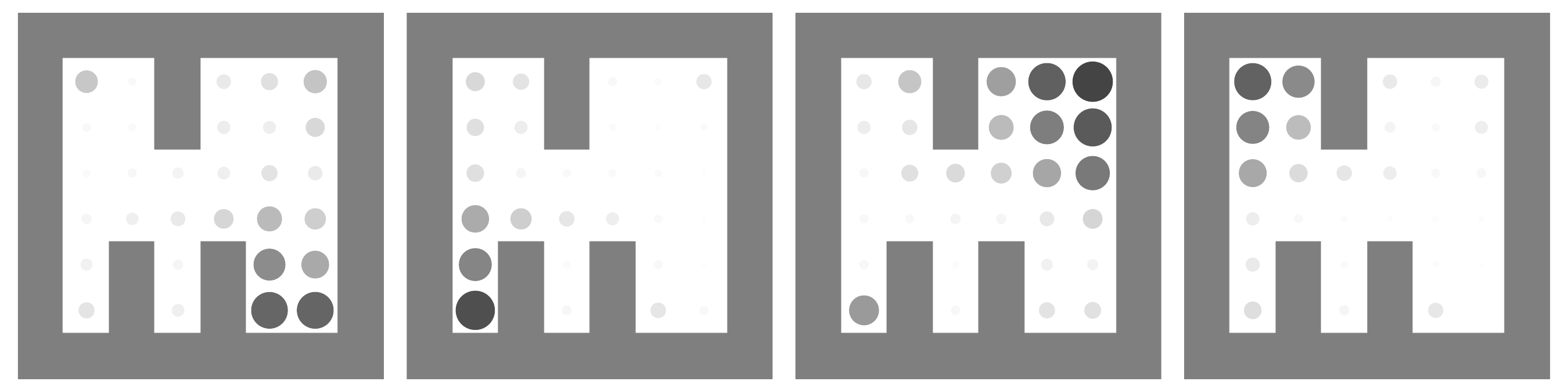}
\end{subfigure}

\vspace*{-1ex}
\caption{\small
	Options learned with \gls{MSOL} on the taxi domain,
	before (top) and after pickup (bottom). 
	The light gray area indicates walls.
	The left plots show the intra-option policies: 
	arrows and colors indicated direction of most likely action, 
	the size indicates its probability. 
	A square indicates the pickup/dropoff action.
	The right plots show the termination policies: 
	intensity and size of the circles indicate termination probability.
	}
% \vspace{-4mm}
\label{fig:taxi}
\end{figure*}

We start with the 2D Moving Bandits environment 
proposed and implemented by \citet{frans2018meta}, 
which is similar to the example in \cref{sec:local_minima}.
There are two randomly sampled, distinguishable, marked positions in the environment. 
In each episode, the agent receives a reward of 1 for each time step it is sufficiently close to the
correct one of both positions, and 0 otherwise.
Which location is rewarded is not signaled in the observation.
The agent can take actions that move it in one of the four cardinal directions.
Each episode lasts 50 steps.

We compare against \gls{MLSH} and \Distral~to highlight challenges that arise in multitask training.
We allow \gls{MLSH} and \gls{MSOL} to learn two options.
During transfer, optimal performance can only be achieved with diverse options that have 
successfully learned to reach {\em different} marked locations.
In \cref{fig:mv} we can see that \gls{MSOL} is able to do so but 
the hard options learned by \gls{MLSH} both learned to reach the \emph{same} goal location,
resulting in only approximately half the optimal return during transfer.
This is exactly the situation outlined in \cref{sec:local_minima} in which learning hard options can
lead to local optima.

\Distral, even with the last action provided as additional input, is not able to quickly utilize the
prior knowledge. 
% Because the locations of the two marked positions are randomly sampled for each episode, 
The last action only conveys meaningful information when taking the goal locations into account:
\Distral~agents need to {\em infer} the intention based on the last action and the relative
goal positions.
While this is possible, in practice the agent was not able to do so, even with a much larger network.
Much longer training ultimately allows \Distral~to perform as well as \gls{MSOL}, denoted by ``\Distral(+action) limit''.
This is not surprising since its posterior is flexible and will therefore eventually be able to
learn any task. However, it is not able to learn transferrable prior knowledge which allows
\emph{fast} training on the new task.
Lastly, \gls{MSOL}(frozen) also outperforms \Distral(+action) and \gls{MLSH},
but performs worse that \gls{MSOL}. %except for extremely short training times.
This highlights the utility of making options soft, i.e. adaptable, during transfer to new tasks. 
It also shows that the advantage of \gls{MSOL} over the other methods lies not only in its flexibility during transfer, but
also during the original learning phase.

% ------------------------------------------------------------------------------
\subsection{TAXI}
\label{sec:taxi}

Next, we use a slightly modified version of the original Taxi domain \citep{dietterich1998maxq} to
show learning of termination functions as well as transfer- and generalization capabilities.
To solve the task, the agent must pick up a passenger 
on one of four possible locations by moving to their
location and executing a special `pickup/drop-off' action. 
Then, the passenger must be dropped off 
at one of the other three locations, 
again using the same action executed at the corresponding location.
The domain has a discrete state space with 30 locations 
arranged on a grid and a flag indicating whether the
passenger was already picked up. 
The observation is a one-hot encoding of the discrete state, excluding passenger- and goal location.
This introduces an information-asymmetry between the task-specific master policy,
and the shared options, allowing them to generalize well \citep{galashov2018information}. 
Walls (see \cref{fig:taxi}) limit the movement of the agent and invalid actions.
% i.e., moving into a wall, result in no change to the state.

We investigate two versions of Taxi. 
In the original, just called \emph{Taxi},
the action space consists of one no-op, 
one `pickup/drop-off' action and four actions to move in all cardinal directions.
In {\em Directional Taxi}, we extend this setup:
the agent faces in one of the cardinal directions and
the available movements are to move forward or 
rotate either clockwise or counter-clockwise.
In both environments the set of tasks $\mathcal{T}$ 
are the 12 different combinations of pickup/drop-off locations.
Episodes last at most 50 steps and
there is a reward of 2 for delivering the passenger 
to its goal and a penalty of -0.1 for each time step.
During training, the agent is initialized to any valid state.
During testing, the agent is always initialized without the passenger on board.

We allow four learnable options in \gls{MLSH} and \gls{MSOL}. 
This necessitates the options to be diverse, 
i.e., one option to reach each of the four pickup/drop-off locations. 
Importantly, it also requires the options to learn to terminate 
when a passenger is picked up.
As one can see in \cref{fig:tn}, 
\gls{MLSH} struggles both with option-diversity and due to its fixed option duration: 
because the starting position is random, the duration until
the option needs to terminate is different between episodes and cannot be captured by one
hyperparameter.
Furthermore, even without correct terminations, one could still learn to solve (at least) four out of
the twelve tasks, leading to an average reward of approximately $-3.2$\footnote{The optimal policy
for a task achieves approximate a return of 0.5 on average whereas the worst possible return is $-5$.}. 
However, \gls{MLSH} is not able to learn diverse enough policies, resulting in worse performance.

\Distral(+action) performs well in the original {\em Taxi} environment, as seen in \cref{fig:tn}.
This is expected since here the last action, moving in a compass direction, 
is a good indicator for the agent's intention,
effectively acting as an optimal ``option'' and inducing temporally extended exploration.
However, in the directional case shown in \cref{fig:ts},
actions rarely indicate intentions,
which makes it much harder for \Distral(+action) to use prior knowledge.
By contrast, \gls{MSOL} performs well in both taxi environments. 
In the directional case, learned \gls{MSOL} options capture temporally
correlated behavior much better than the last action in \Distral.
% Comparing its performance with \gls{MSOL}(frozen) shows the utility of adaptable soft options during transfer.

%\cref{fig:taxi}, which visualizes the options learned by \gls{MSOL}, 
%shows that it successfully learns useful movement primitives 
%and termination functions.
\cref{fig:taxi} demonstrates that the options learned by \gls{MSOL}
learn movement and termination policies that make intuitive sense.
Note that the same soft option represents different behavior 
depending on whether it already picked up the passenger,
%This is expected 
as this behavior does not need 
to terminate the current option on three of the 12 tasks.

% ------------------------------------------------------------------------------
\subsection{OUT-OF-DISTRIBUTION TASKS}
\label{sec:out_of_dist}
\begin{figure*}[t]
    \centering
    \begin{subfigure}[b]{0.32\textwidth}
      \includegraphics[width=1\linewidth]{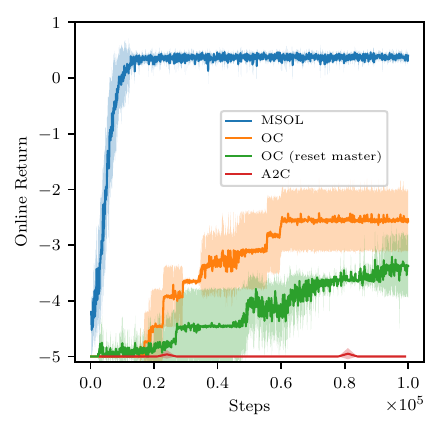}
      % \vspace*{-6mm}
      \caption{\small Taxi: Generalization}%\vspace*{-1.5ex}
      \label{fig:generalization}
    \end{subfigure}
    \begin{subfigure}[b]{0.32\textwidth}
      \includegraphics[width=1\linewidth]{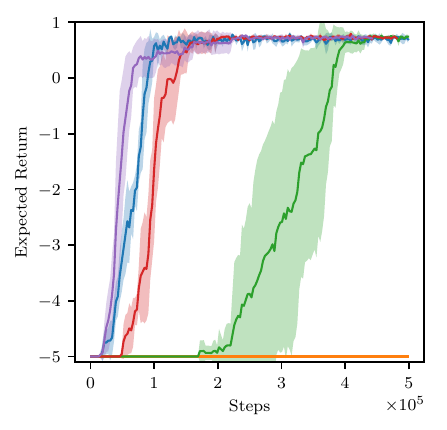}
      % \vspace*{-6mm}
      \caption{\small Taxi (small): Adaptation}%\vspace*{-1.5ex}
      \label{fig:change_in_task_small}
    \end{subfigure}
    \begin{subfigure}[b]{0.32\textwidth}
      \includegraphics[width=1\linewidth]{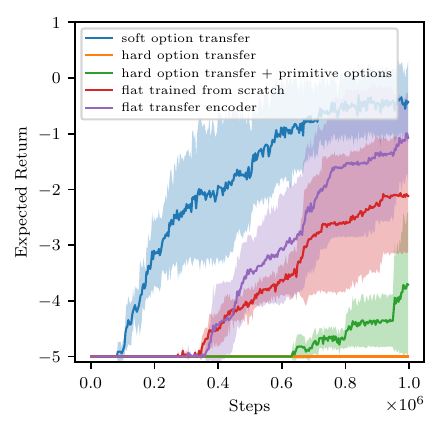}
      % \vspace*{-6mm}
      \caption{\small Taxi (large): Adaptation}%\vspace*{-1.5ex}
      \label{fig:change_in_task_large}
    \end{subfigure}
    \vspace*{-1ex}
    \caption{\small
    % We investigate the utility of options when transferred to unseen and out of distribution tasks. 
    We compare \gls{MSOL} against \glsentryfull{OC}, hard options and flat policies trained from scratch or
    with a pre-trained encoder. For a fair comparison, the soft option prior is identical to the
    hard option in these experiments.
    \emph{Left:} Since the options in \gls{OC} are not task-agnostic, they fail to generalize to previously
    unseen tasks.
    \emph{Middle and right:} Transfer performance of options 
    to environments in which the pickup and dropoff locations where shifted, making the options
    misspecified.
    Only soft options provide utility over flat policies in this setting.
    The middle figure shows results on a small grid in which exploration is simple, whereas the
    right figure shows that transfer learning can accelerate exploration especially on larger tasks.
    }
    \label{fig:performance2}
    \vspace{-2mm}
  \end{figure*}

In this section, we show how learning soft options can help with transfer to unseen tasks.
In \cref{fig:generalization} we show learning on four tasks from $\mathcal{T}$ using options that
were trained on the remaining eight, comparing against \glsentryfull{A2C} \citep{mnih2016asynchronous} and \glsentryfull{OC} \citep{bacon2017option}.
Note that in \gls{OC}, there is no information-asymmetry:
the same networks are shared across all tasks 
and provided with a task-id as additional input, including
to the option-policies. This prevents \gls{OC} from generalizing well to unseen tasks.
On the other hand, withholding the task-information 
would be similar to \gls{MLSH}, which
we already showed to struggle with local minima.
The strong performance of \gls{MSOL} shows that %it helps 
%to train information-asymmetric options to generalize to new task.
information-asymmetric options help to generalize to previously unseen tasks.

We also investigate the utility of flexible soft options under a shift of the task distribution: 
in \cref{fig:change_in_task_small,fig:change_in_task_large} we show
learning performance on twelve \emph{modified} tasks in which the pickup/dropoff locations where moved by
one cell while the options were trained with the original locations. 
While the results in \cref{fig:change_in_task_small} use a smaller grid,
\cref{fig:change_in_task_large} shows the results for a larger grid in which exploration is more difficult.
As expected, hard options are not able to solve this task for either grid-size. 
Moreover, while combining hard options with primitive actions allows the tasks to be solved
eventually, it performs worse than training a new, flat policy from scratch. 
The finding that access to misspecified, hard options can actually \emph{hurt} exploration is consistent
with previous literature \citep{jong2008utility}.
On the other hand, \gls{MSOL} is able to quickly learn on this new task by adapting the previously
learned options. 

Note that on the small grid in which exploration is easy, our hierarchical method performs
similar to a flat policy. On the larger grid exploration becomes more challenging and 
\gls{MSOL} learns significantly faster, highlighting how transfer learning can improve exploration.
More results can be found in \cref{sec:ap:taxi}.
\vspace*{-1.0ex}
\section{DISCUSSION}
\vspace*{-1ex}

% M S O L brings multiple advantages: It stabilizes end-to-end multitask training, removing the need
% for complex training schedules like in (Frans et al., 2018). Furthermore, it allows for coordination
% between master policies, avoiding local minima of the type described in Figure 1(a). Lastly, it
% allows fine-tuning of options on new tasks at test-time. 

% In this work, we propose Multitask Soft Option Learning (\gls{MSOL}).
%, derived from the \gls{PAI} formulation of \gls{RL}.
%
Multitask Soft Option Learning (\gls{MSOL}) proposes 
reformulating options using the perspective of prior and posterior distributions. This
offers several key advantages.
First, during transfer, it allows us to distinguish between fixed, and therefore knowledge-preserving option {\em
priors}, and flexible option {\em posteriors} that can adjust to the reward structure of the task at
hand.
This effects a similar speed-up in learning as the original options framework, while avoiding sub-optimal performance when the available options are not perfectly aligned to the task.
Second, utilizing this `soft' version of options in a multitask learning setup increases
optimization stability and removes the need for complex training schedules.
Furthermore, this framework naturally allows master policies to coordinate across tasks and
avoid local minima of insufficient option diversity.
It also allows for autonomously learning option-termination policies, a very challenging task which
is often avoided by fixing option durations manually.

Lastly, using this formulation also allows inclusion of prior information in a principled manner without imposing too rigid a structure on the resulting hierarchy.
We utilize this advantage to explicitly incorporate the bias that good options should be temporally extended.
In future research, other types of information can be explored.
As an example, one could investigate sets of tasks which would benefit from a learned master prior, 
like walking on different types of terrain.
%

%%% Local Variables:
%%% mode: latex
%%% TeX-master: "../main"
%%% End:

\vspace*{-1.0ex}
\section{ACKNOWLEDGEMENTS}
\vspace*{-1ex}
MI is supported by the AIMS EPSRC CDT.
NS, AG, NN were funded by ERC grants ERC-2012-AdG 321162-HELIOS and Seebibyte EP/M013774/1, and EPSRC/MURI grant EP/N019474/1.
SW is supported by ERC under the Horizon 2020 research and innovation programme (grant agreement number 637713).

\newpage

\bibliographystyle{abbrvnat}
\bibliography{refs}

\begin{thebibliography}{32}
\providecommand{\natexlab}[1]{#1}
\providecommand{\url}[1]{\texttt{#1}}
\expandafter\ifx\csname urlstyle\endcsname\relax
  \providecommand{\doi}[1]{doi: #1}\else
  \providecommand{\doi}{doi: \begingroup \urlstyle{rm}\Url}\fi

\bibitem[Andreas et~al.(2017)Andreas, Klein, and Levine]{andreas2017modular}
J.~Andreas, D.~Klein, and S.~Levine.
\newblock Modular multitask reinforcement learning with policy sketches.
\newblock In \emph{ICML}, 2017.

\bibitem[Bacon et~al.(2017)Bacon, Harb, and Precup]{bacon2017option}
P.-L. Bacon, J.~Harb, and D.~Precup.
\newblock The option-critic architecture.
\newblock In \emph{AAAI}, 2017.

\bibitem[Bengio et~al.(2009)Bengio, Louradour, Collobert, and
  Weston]{bengio2009curriculum}
Y.~Bengio, J.~Louradour, R.~Collobert, and J.~Weston.
\newblock Curriculum learning.
\newblock ACM, 2009.

\bibitem[Botvinick et~al.(2009)Botvinick, Niv, and
  Barto]{botvinick2009hierarchically}
M.~M. Botvinick, Y.~Niv, and A.~C. Barto.
\newblock Hierarchically organized behavior and its neural foundations: a
  reinforcement learning perspective.
\newblock \emph{Cognition}, \penalty0 (3), 2009.

\bibitem[Daniel et~al.(2012)Daniel, Neumann, and
  Peters]{daniel2012hierarchical}
C.~Daniel, G.~Neumann, and J.~Peters.
\newblock Hierarchical relative entropy policy search.
\newblock In \emph{Artificial Intelligence and Statistics}, 2012.

\bibitem[Daniel et~al.(2016)Daniel, Van~Hoof, Peters, and
  Neumann]{daniel2016probabilistic}
C.~Daniel, H.~Van~Hoof, J.~Peters, and G.~Neumann.
\newblock Probabilistic inference for determining options in reinforcement
  learning.
\newblock \emph{Machine Learning}, 2016.

\bibitem[Dietterich(1998)]{dietterich1998maxq}
T.~G. Dietterich.
\newblock The {MAXQ} method for hierarchical reinforcement learning.
\newblock In \emph{ICML}, 1998.

\bibitem[Fox et~al.(2016)Fox, Moshkovitz, and Tishby]{fox2016principled}
R.~Fox, M.~Moshkovitz, and N.~Tishby.
\newblock Principled option learning in markov decision processes.
\newblock \emph{arXiv:1609.05524}, 2016.

\bibitem[Frans et~al.(2018)Frans, Ho, Chen, Abbeel, and
  Schulman]{frans2018meta}
K.~Frans, J.~Ho, X.~Chen, P.~Abbeel, and J.~Schulman.
\newblock Meta learning shared hierarchies.
\newblock In \emph{ICLR}, 2018.

\bibitem[Galashov et~al.(2019)Galashov, Jayakumar, Hasenclever, Tirumala,
  Schwarz, Desjardins, Czarnecki, Teh, Pascanu, and
  Heess]{galashov2018information}
A.~Galashov, S.~Jayakumar, L.~Hasenclever, D.~Tirumala, J.~Schwarz,
  G.~Desjardins, W.~M. Czarnecki, Y.~W. Teh, R.~Pascanu, and N.~Heess.
\newblock Information asymmetry in {KL}-regularized {RL}.
\newblock In \emph{ICLR}, 2019.

\bibitem[Goyal et~al.(2019)Goyal, Islam, Strouse, Ahmed, Larochelle, Botvinick,
  Levine, and Bengio]{goyal2018transfer}
A.~Goyal, R.~Islam, D.~Strouse, Z.~Ahmed, H.~Larochelle, M.~Botvinick,
  S.~Levine, and Y.~Bengio.
\newblock Transfer and exploration via the information bottleneck.
\newblock In \emph{ICLR}, 2019.

\bibitem[Gregor et~al.(2016)Gregor, Rezende, and
  Wierstra]{gregor2016variational}
K.~Gregor, D.~J. Rezende, and D.~Wierstra.
\newblock Variational intrinsic control.
\newblock \emph{arXiv:1611.07507}, 2016.

\bibitem[Haarnoja et~al.(2018)Haarnoja, Hartikainen, Abbeel, and
  Levine]{haarnoja2018latent}
T.~Haarnoja, K.~Hartikainen, P.~Abbeel, and S.~Levine.
\newblock Latent space policies for hierarchical reinforcement learning.
\newblock \emph{arXiv:1804.02808}, 2018.

\bibitem[Harb et~al.(2017)Harb, Bacon, Klissarov, and Precup]{harb2017waiting}
J.~Harb, P.-L. Bacon, M.~Klissarov, and D.~Precup.
\newblock When waiting is not an option: Learning options with a deliberation
  cost.
\newblock \emph{arXiv:1709.04571}, 2017.

\bibitem[Harutyunyan et~al.(2019)Harutyunyan, Dabney, Borsa, Heess, Munos, and
  Precup]{harutyunyan2019termination}
A.~Harutyunyan, W.~Dabney, D.~Borsa, N.~Heess, R.~Munos, and D.~Precup.
\newblock The termination critic.
\newblock \emph{arXiv:1902.09996}, 2019.

\bibitem[Hausman et~al.(2018)Hausman, Springenberg, Wang, Heess, and
  Riedmiller]{hausman2018learning}
K.~Hausman, J.~T. Springenberg, Z.~Wang, N.~Heess, and M.~Riedmiller.
\newblock Learning an embedding space for transferable robot skills.
\newblock In \emph{ICLR}, 2018.

\bibitem[Heess et~al.(2016)Heess, Wayne, Tassa, Lillicrap, Riedmiller, and
  Silver]{heess2016learning}
N.~Heess, G.~Wayne, Y.~Tassa, T.~Lillicrap, M.~Riedmiller, and D.~Silver.
\newblock Learning and transfer of modulated locomotor controllers.
\newblock \emph{arXiv:1610.05182}, 2016.

\bibitem[Jong et~al.(2008)Jong, Hester, and Stone]{jong2008utility}
N.~K. Jong, T.~Hester, and P.~Stone.
\newblock The utility of temporal abstraction in reinforcement learning.
\newblock In \emph{AAMAS}. International Foundation for Autonomous Agents and
  Multiagent Systems, 2008.

\bibitem[Levine(2018)]{levine2018reinforcement}
S.~Levine.
\newblock Reinforcement learning and control as probabilistic inference:
  Tutorial and review.
\newblock \emph{arXiv:1805.00909}, 2018.

\bibitem[McGovern and Barto(2001)]{mcgovern2001automatic}
A.~McGovern and A.~G. Barto.
\newblock Automatic discovery of subgoals in reinforcement learning using
  diverse density.
\newblock In \emph{ICML}, 2001.

\bibitem[Mnih et~al.(2016)Mnih, Badia, Mirza, Graves, Lillicrap, Harley,
  Silver, and Kavukcuoglu]{mnih2016asynchronous}
V.~Mnih, A.~P. Badia, M.~Mirza, A.~Graves, T.~Lillicrap, T.~Harley, D.~Silver,
  and K.~Kavukcuoglu.
\newblock Asynchronous methods for deep reinforcement learning.
\newblock In \emph{ICML}, 2016.

\bibitem[Nachum et~al.(2018)Nachum, Gu, Lee, and Levine]{nachum2018data}
O.~Nachum, S.~Gu, H.~Lee, and S.~Levine.
\newblock Data-efficient hierarchical reinforcement learning.
\newblock \emph{arXiv:1805.08296}, 2018.

\bibitem[Ng et~al.(1999)Ng, Harada, and Russell]{ng1999policy}
A.~Y. Ng, D.~Harada, and S.~Russell.
\newblock Policy invariance under reward transformations: Theory and
  application to reward shaping.
\newblock In \emph{ICML}, 1999.

\bibitem[Schulman et~al.(2017)Schulman, Wolski, Dhariwal, Radford, and
  Klimov]{schulman2017proximal}
J.~Schulman, F.~Wolski, P.~Dhariwal, A.~Radford, and O.~Klimov.
\newblock Proximal policy optimization algorithms.
\newblock \emph{arXiv:1707.06347}, 2017.

\bibitem[Sutton et~al.(1999)Sutton, Precup, and Singh]{sutton1999between}
R.~S. Sutton, D.~Precup, and S.~Singh.
\newblock Between mdps and semi-mdps: A framework for temporal abstraction in
  reinforcement learning.
\newblock \emph{Artificial intelligence}, \penalty0 (1-2), 1999.

\bibitem[Teh et~al.(2017)Teh, Bapst, Czarnecki, Quan, Kirkpatrick, Hadsell,
  Heess, and Pascanu]{teh2017distral}
Y.~Teh, V.~Bapst, W.~M. Czarnecki, J.~Quan, J.~Kirkpatrick, R.~Hadsell,
  N.~Heess, and R.~Pascanu.
\newblock Distral: Robust multitask reinforcement learning.
\newblock In \emph{NeurIPS}, 2017.

\bibitem[Tessler et~al.(2017)Tessler, Givony, Zahavy, Mankowitz, and
  Mannor]{tessler2017deep}
C.~Tessler, S.~Givony, T.~Zahavy, D.~J. Mankowitz, and S.~Mannor.
\newblock A deep hierarchical approach to lifelong learning in minecraft.
\newblock In \emph{AAAI}, 2017.

\bibitem[Thrun and Schwartz(1995)]{thrun1995finding}
S.~Thrun and A.~Schwartz.
\newblock Finding structure in reinforcement learning.
\newblock In \emph{NeurIPS}, 1995.

\bibitem[Tirumala et~al.(2019)Tirumala, Noh, Galashov, Hasenclever, Ahuja,
  Wayne, Pascanu, Teh, and Heess]{tirumala2019exploiting}
D.~Tirumala, H.~Noh, A.~Galashov, L.~Hasenclever, A.~Ahuja, G.~Wayne,
  R.~Pascanu, Y.~W. Teh, and N.~Heess.
\newblock Exploiting hierarchy for learning and transfer in kl-regularized rl.
\newblock \emph{arXiv:1903.07438}, 2019.

\bibitem[Todorov(2008)]{todorov2008general}
E.~Todorov.
\newblock General duality between optimal control and estimation.
\newblock IEEE, 2008.

\bibitem[Vezhnevets et~al.(2017)Vezhnevets, Osindero, Schaul, Heess, Jaderberg,
  Silver, and Kavukcuoglu]{vezhnevets2017feudal}
A.~S. Vezhnevets, S.~Osindero, T.~Schaul, N.~Heess, M.~Jaderberg, D.~Silver,
  and K.~Kavukcuoglu.
\newblock Feudal networks for hierarchical reinforcement learning.
\newblock In \emph{ICML}, 2017.

\bibitem[Wang et~al.(2016)Wang, Kurth-Nelson, Tirumala, Soyer, Leibo, Munos,
  Blundell, Kumaran, and Botvinick]{wang2016learning}
J.~X. Wang, Z.~Kurth-Nelson, D.~Tirumala, H.~Soyer, J.~Z. Leibo, R.~Munos,
  C.~Blundell, D.~Kumaran, and M.~Botvinick.
\newblock Learning to reinforcement learn.
\newblock \emph{arXiv:1611.05763}, 2016.

\end{thebibliography}

\appendix

\clearpage
\appendix
\onecolumn

\part*{APPENDIX}

% ==============================================================================

\section{PSEUDO CODE}

\begin{algorithm}[h] 
	\caption{Pseudo-Code for MSOL} 
	\textbf{Input} Number $m$ of options to learn, $n_i$ different training tasks $\mathrm{env}_i$, \emph{fixed}
  termination prior $p^T(b_t) = (1-\alpha)^{b_t} \alpha^{1-b_t}$ and \emph{fixed} master prior $p^H(z_t | z_{t-1}, b_t) = (1 - b_t) \, \delta(z_t - z_{t-1}) \,+\, b_t {\textstyle\frac{1}{m}}$\\
    \textbf{Initialize once:} \emph{Learnable} termination prior $p_\theta^T(b_t|s_t, z_{t-1})$, intra-option prior
    $p_\theta^L(a_t|s_t,z_t)$ \\
    \textbf{Initialize for each task $i$:} Lernable termination posterior $q_{\phi_i}^T(b_t|s_t, z_{t-1})$,
    master policy $q_{\phi_i}^H(z_t|s_t,z_{t-1}, b_t)$, intra-option policy $q_{\phi_i}^L(a_t|s_t,z_t)$\\
    \emph{// Note that this leads to a total of $m$ intra option priors for the $m$ different values
    of $z\in\{1\dots m\}$}\\
    \emph{// and a total of $m\times n_i$ intra option posteriors.} \\
    \BlankLine
    \BlankLine
    \While{\upshape not converged} 
    {
      \emph{// Collect data}\\
      \For {\upshape each task $i$}{
        \If{\upshape beginning of episode}{
          $s_0 \leftarrow \mathrm{env.reset}()$\\
          $b_0 \leftarrow 1$ \emph{// This allows $q^H$ to sample a new option $z_0$.} \\
        }
        \Else{
        $b_t\sim q_{\phi_i}^T(b_t|s_t, z_{t-1})$ \\
        }

        \BlankLine
        $z_t\sim q_{\phi_i}^H(z_t|s_t,z_{t-1}, b_t)$ \\
        $a_t\sim q_{\phi_i}^L(a_t|s_t,z_t)$ \\
        $s_{t+1}, r_{t} \sim \mathrm{env}_i(a_t)$ \\
        \BlankLine
        \emph{// Compute regularized reward (eq. \cref{eq:full_reg_reward}); note that we use the fixed priors here}
          $R_{t}^\text{reg} \leftarrow r_i(s_t,a_t)
            - \beta \ln{\textstyle\frac{q^H_{\phi_i}(z_t|s_t, z_{t-1}, b_t)}%
                {p^H(z_t|z_{t-1}, b_t)}} 
           - \beta \ln{\textstyle\frac{q^L_{\phi_i}(a_t|s_t, z_t)}%
              {p^L_\theta(a_t|s_t, z_t)}}
          - \beta \ln{\textstyle\frac{q^T_{\phi_i}(b_t|s_t, z_{t-1})}%
              {p^T(b_t)}}$
              % {p_\theta^T\!(b_t|s_t,z_{t-1})}}

        Add $s_t, s_{t+1}, r_{t}, R_t^\text{reg}$ to $\mathcal{D}_i$ \\
        
      }
      \BlankLine
      \BlankLine
      \emph{// Update parameters $\phi_i$} \\
      \For {\upshape each task $i$}{
        Update $\phi_i$ using A2C or PPO on $\mathcal{D}_i$ as desribed in \cref{sec:gradient-updates}. 
        Note that for PPO, $R_t^\text{reg}$ needs to be re-computed and updated between gradient
        updates to $\phi_i$ as the regularization terms change.
      }
      \BlankLine
      \BlankLine
      \emph{// Update parameters $\theta$} \\
      % Combine $\mathcal{D} \leftarrow \{\mathcal{D}_i\}_{i=1\dots n_i}$ \\ 
      Update $\theta$ to minimize $\sum_i\mathbb{E}_{\mathcal{D}_i} \left[
      \KL{q^L_{\phi_i}(a_t|s_t, z_t)}{p_\theta^L(a_t|s_t,z_t)} + 
      \KL{\hat q_{\phi_i}(b=1|s_t, z_{t-1})} {p_\theta^T(b_t|s_t, z_{t-1})} \right]$ \\
      Here, $\hat q_{\phi_i}(b=1|s_t, z_{t-1}) = \sum_{z_t\neq z_{t-1}} q^H_{\phi_i}(z_t
      |s_t, z_{t-1},b_t = 1)$, i.e. instead of distilling the average of $q^T_{\phi_i}$ into $p^T_\theta$, we
      distill whether the \emph{master policy} $q^H_{\phi_i}$ would have changed the option $z_t$ if it had
      the chance (i.e. if $b_t=1$). Since $q^T_{\phi_i}$ is regularized to be similar to the
      \emph{fixed} $p^T$, this approach allows us to learn a termination prior
      $p^T_\theta$ which is less influenced by our manually specified prior $p^T$, and more by what
      is needed for the task. \\
    }
\end{algorithm}

\section{MSOL TRAINING DETAILS}
\label{sec:training_details}

% ------------------------------------------------------------------------------
\subsection{OPTIMIZATION}
\label{sec:gradient-updates}
\def\Rr{R^\text{reg}}

Even though $R^{\text{reg}}_i$ depends on $\phi_i$, its gradient w.r.t.\ $\phi_i$ vanishes.\footnote{
	$\int p(x) \, \nabla\ln p(x) \, dx = \int \nabla p(x) \, dx = \nabla \int p(x) \, dx = 0$.
}
Consequently, we can treat the regularized reward as a classical \gls{RL} reward and use any \gls{RL} algorithm to find the optimal hierarchical policy parameters $\phi_i$. 
In the following, we explain how to adapt A2C \citep{mnih2016asynchronous} to soft options.
The extension to PPO \citep{schulman2017proximal} is straightforward.\footnote{However,
for \gls{PAI} frameworks like ours, unlike in the original PPO implementation, the advantage function must be updated after each
epoch.}

The joint posterior policy in (\ref{eq:policy_posterior}) depends on the 
current state $s_t$ and the previously selected option $z_{t-1}$.
The expected sum of regularized future rewards of task $i$,
the value function $V_i$,
must therefore also condition on this pair:
\begin{equation}
	V_i(s_t, z_{t-1}) \;:=\;
		\E_{\tau \sim q} \Big[ \, {\textstyle\sum\limits_{t'=t}^T}
				\gamma^{t'-t} \Rr_{i,t'} \,\Big|\, s_t, z_{t-1} \Big] \,.
\end{equation}  
As $V_i(s_t, z_{t-1})$ cannot be directly observed,
we approximate it with a parametrized model $V_{\phi_i}(s_t, z_{t-1})$.
The $k$-step advantage estimation at time $t$ of trajectory $\tau$ is given by
\begin{equation}
	A_{\phi_i}(\tau_{t:(t+k)}) 
	\quad := \quad
	\smallsum{j=0}{k-1}\gamma^j\Rr_{t+j} 
		+ \gamma^k V_{\phi_i}^-(s_{t+k}, z_{t+k-1}) 
		- V_{\phi_i}(s_t, z_{t-1}) \,,
\end{equation}
where the superscript `\(-\)' indicates treating the term as a constant.
The approximate value function $V_{\phi_i}$ 
can be optimized towards its bootstrapped $k$-step target 
by minimizing 
$\mathcal{L}_V(\phi_i,\tau_{1:T}) := 
\sum_{t=1}^T (A_{\phi_i}\!(\tau_{t:(t+k)}))^2$.
As per A2C, $k\in[1\dots n_s]$ depending on the state \citep{mnih2016asynchronous}.
The corresponding policy gradient loss is
\vspace{0mm}
\begin{equation} \nonumber
	\L_A(\phi_i, \tau_{1:T}) := {\textstyle \sum\limits_{t=1}^{T}} 
	A^{-}_{\phi_i}\!(\tau_{t:(t+k)})   
	\ln q_{\phi_i}\!(a_t, z_t, b_t|s_t, z_{t-1}) \,.
\end{equation}
\vspace{-3mm}

The gradient w.r.t.~the prior parameters $\theta$ 
is\footnote{Here we ignore $\beta$ as it is folded into $\lambda_P$ later.}
\begin{equation}
	\nabla_\theta  \L_P(\theta,\tau_{1:T}, \tilde b_{1:T}) 
	\quad := \quad
 	-\! {\textstyle\sum\limits_{t=1}^T} \Big(
 	\nabla_\theta \ln p^L_\theta(a_t|s_t, z_t)
 	+ \nabla_\theta \ln p^T_\theta(\tilde{b}_t|s_t, z_{t-1}) \Big) \,,
\end{equation}
where $\tilde{b}_t=\delta_{z_{t-1}}(z_t')$ and $z_t'\sim q^H(z_t'|s_t, z_{t-1}, b_t=1)$.
To encourage exploration in all policies of the hierarchy,
we also include an entropy maximization loss:
\vspace{-2mm}
\begin{equation}
	\L_H(\phi_i, \tau_{1:T}) 
	\quad := \quad  
	{\sum\limits_{t=1}^T} \Big(
		\ln q^H_{\phi_i}\!(z_t | s_t, z_{t-1}, b_t) 
		+ \ln q^L_{\phi_i}(a_t|s_t,z_t) 
		+ \ln q^T_{\phi_i}(b_t|s_t, z_{t-1}) \Big) \,.
\end{equation} 
Note that term \circled{1} in \eqref{eq:full_reg_reward} 
already encourages maximizing
$\L_H(\phi_i, \tau)$ for the master policy, 
since we chose a uniform prior $p^H(z_t|b_t=1)$.
As both terms serve the same purpose, we are free to drop either one of them.
In our experiments, we chose to drop the term for $q^H$ in $R^\text{reg}_t$, 
which proved slightly more stable to optimize that the alternative.

We can optimize all parameters jointly with
a combined loss over all tasks $i$, 
based on sampled trajectories $\tau^i := \tau^i_{1:T} \sim q_{\phi_i}$
and corresponding sampled values of $\tilde b^i := \tilde{b}_{1:T}^i$:
\vspace{-2mm}
\begin{equation} \nonumber \label{eq:final_loss}
  \L(\{\phi_i\}, \theta, \{\tau^i\}, \{\tilde b^i\}) 
  = \sum_{i=1}^{n} \Big( 
  		\L_A(\phi_i, \tau^i)  + \lambda_V \L_V(\phi_i, \tau^i)  
  		+ \lambda_P \L_P(\theta, \tau^i, \tilde b^i)  
  		+ \lambda_H \L_H(\phi_i, \tau^i)  
  \Big) \,. 
\end{equation}

\subsection{TRAINING SCHEDULE}

For faster training, it is important to prevent the master policies $q^H$ from converging too
quickly to allow sufficient updating of all options. On the other hand, a lower exploration rate leads to more clearly defined options. We
consequently anneal the exploration bonus $\lambda_H$ with a linear schedule during training.

Similarly, a high value of $\beta$ leads to better options but can prevent finding the extrinsic
reward $r_i(s_t, a_t)$ early on in training. Consequently, we increase $\beta$ over the course of
training, also using a linear schedule.

% ==============================================================================
\section{ARCHITECTURE}
\label{sec:architecture}

All policies and value functions share the same encoder network with two fully connected hidden layers of size 64 for the Moving
Bandits environment and three hidden layers of sizes 512, 256, and 512 for the Taxi environments.
Distral was tested with both model sizes on the Moving Bandits task to make sure
that limited capacity is not the problem. Both models resulted in similar performance, the results
shown in the paper are for the larger model. 
Master-policies, as well as all prior- and posterior policies and value functions consist of only
one layer which takes the latent embedding produced by the encoder as input.
Furthermore, the encoder is shared across tasks, allowing for much faster training since
observations can be batched together.

Options are specified as an additional one-hot encoded input to the corresponding network that is
passed through a single 128 dimensional fully connected layer and concatenated to the state
embedding before the last hidden layer.
We implement the single-column architecture of Distral as a hierarchical policy with just one option and
with a modified loss function that does not include terms for the master and termination policies.
Our implementation builds on the A2C/PPO implementation by, and we use the
implementation for \gls{MLSH} that is provided by the authors (\url{https://github.com/openai/mlsh}).

\begin{figure*}[b!]
  \centering
  \begin{subfigure}[b]{0.32\textwidth}
    \includegraphics[width=1\linewidth]{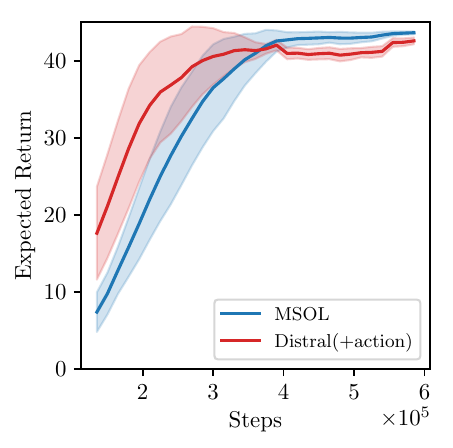}
    \caption{\small Moving Bandits}\vspace*{-1.5ex}
  \end{subfigure}
  \begin{subfigure}[b]{0.32\textwidth}
    \includegraphics[width=1\linewidth]{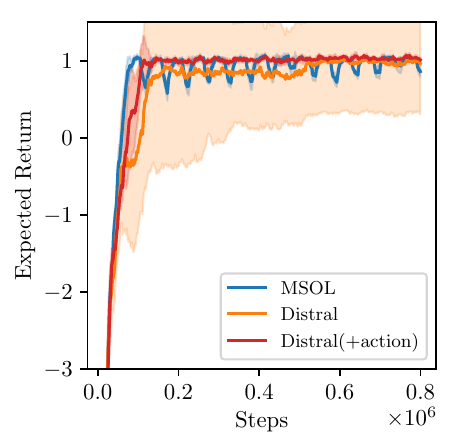}
    \caption{\small Taxi}\vspace*{-1.5ex}
  \end{subfigure}
  \begin{subfigure}[b]{0.32\textwidth}
    \includegraphics[width=1\linewidth]{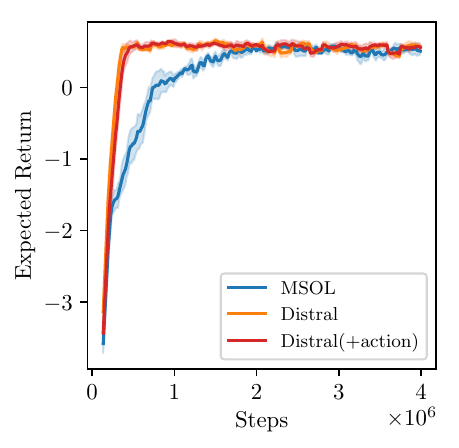}
    \caption{\small Directional Taxi}\vspace*{-1.5ex}
  \end{subfigure}
  % %
  % \begin{subfigure}[b]{0.24\textwidth}
  %   \includegraphics[width=1\linewidth]{images/exp_training/Swimmer}
  %   \caption{\small Swimmer}\vspace*{-1.5ex}
  % \end{subfigure}
  % %
  \caption{Performance during training phase. Note that \gls{MSOL} and \gls{MSOL}(frozen) share the
  same training as they only differ during testing. Further, note that the highest achievable
  performance for Taxi and Directional Taxi is higher during training as they can be initialized
  closer to the final goal (i.e. with the passenger on board).}
  \label{fig:training}
\end{figure*}

% ==============================================================================
\section{HYPER-PARAMETERS AND ADDITIONAL ENVIRONMENT DETAILS}
\label{sec:hyper-params}

We use $2\lambda_V=\lambda_A=\lambda_P=1$ in all experiments.
Furthermore, we train on all tasks from the task distribution, regularly resetting individual tasks
by resetting the corresponding master and re-initializing the posterior policies.
Optimizing $\beta$ for \gls{MSOL} and Distral was done over $\{0.01, 0.02, 0.04, 0.1, 0.2, 0.4\}$.
We use $\gamma=0.95$ for Moving Bandits and Taxi.

\subsection{MOVING BANDITS}

For \gls{MLSH}, we use the original hyper-parameters \citep{frans2018meta}.
The duration of each option is fixed to 10.
The required warm-up duration is set to 9 and the training duration set to 1.
We also use 30 parallel environments split between 10 tasks.
This and the training duration are the main differences to the original paper.
Originally, \gls{MLSH} was trained on 120 parallel environments which we were unable to do due to
hardware constraints. Training is done over 6 million frames per task.

For \gls{MSOL} and Distral we use the same number of 10 tasks and 30 processes. The duration of options are learned
and we do not require a warm-up period. We set the learning rate to $0.01$ and $\beta=0.2$, $\alpha=0.95$, $\lambda_H=0.05$.
Training is done over 0.6 million frames per task.
For Distral we use $\beta=0.04$, $\lambda_H=0.05$ and also 0.6 million frames per task.

\subsection{TAXI}
\label{sec:ap:taxi}

\begin{figure*}[t!]
  \centering
  \begin{subfigure}[b]{0.45\textwidth}
    \includegraphics[width=1\linewidth]{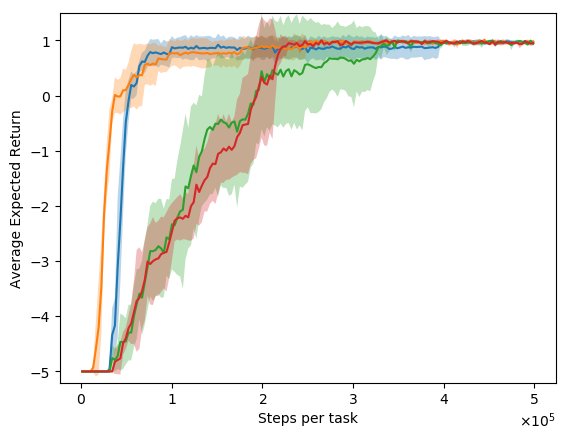}
    % \caption{\small Grid of size 8x8}
    % \label{fig:change_in_task_8x8}
  \end{subfigure}
  \begin{subfigure}[b]{0.45\textwidth}
    \includegraphics[width=1\linewidth]{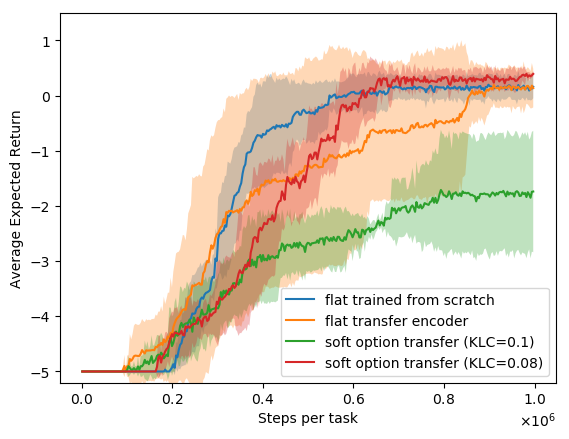}
    % \caption{\small Grid of size 10x10}
    % \label{fig:change_in_task_10x10}
  \end{subfigure}
  \caption{\small
  Results on a `further modified' taxi environment in which the goal locations at test time were
  shifted compared to training, making the learned options misspecified, similar to
  \cref{fig:change_in_task_small,fig:change_in_task_large}.
  Here, the goal locations were shifted further, making the options more misspecified.
  \emph{Left:} Results on a `small' 8x8 grid. \emph{Right:} Results on a `large' 10x10 grid.
  }
  \label{fig:further_modified_results}
\end{figure*}

For \gls{MSOL} we anneal $\beta$ from 0.02 to 0.1 and $\lambda_H$ from 0.1 to 0.05. For Distral we
use $\beta=0.04$. We use 3 processes per task to collect experience for a batch size of 15 per task.
Training is done over 1.4 million frames per task for {\em Taxi} and 4 million
frames per task for {\em Directional Taxi}. 
MLSH was trained on 0.6 million frames for {\em
Taxi} as due to it's long runtime of several days, using more frames was infeasible. Training was
already converged. 

\paragraph{Tasks further out of distribution}
In \cref{fig:further_modified_results} we provided additional results for \cref{sec:out_of_dist}.
We show the performance on \emph{further modified} environments, for which the goal locations were
moved by a second block from the original location for which the options were trained.
We only compare soft options with flat policies trained from scratch, as we already showed in
\cref{sec:out_of_dist} that hard options are unable to cope well with goal modifications.

As expected, the flat policy trained from scratch performs similarly as before, as the moved goal
location does not impact it much. 
On the other hand, using a pre-trained encoder performs slightly worse. 
On the smaller task (left figure) the options are too misspecified to be competitive, despite being soft. 
For the larger grid (right figure) and for a sufficienlty small value of $\beta$ (`KLC'), the options, despite misspecification, are
still competitive.

Consequently, while there is no hard limitation of our approach for appropriately chosen $\beta$, if the target task
is too different from the source task, it will be faster to learn a new policy from scratch. 
Which algorithm trains faster depends mainly on the difficulty of exploration in the target task.
Hard exploration makes options more useful compared to a new, flat policy, even if the options are
misspecified. However, the more misspecified the options are, the smaller the advantage. If target
and source task are too different, very little positive transfer
can be expected, and learning a new (flat) policy becomes more efficient.

\end{document}